\documentclass[conference]{IEEEtran}
\IEEEoverridecommandlockouts
\usepackage{cite}
\usepackage{amsmath,amssymb,amsfonts}
\usepackage{algorithmic}
\usepackage{graphicx}
\usepackage{textcomp}
\usepackage{xcolor}
\usepackage{multirow}
\usepackage{url}
\usepackage{soul}
\usepackage{hyperref}

\def\BibTeX{{\rm B\kern-.05em{\sc i\kern-.025em b}\kern-.08em
    T\kern-.1667em\lower.7ex\hbox{E}\kern-.125emX}}
\begin{document}


\title{On the Reliability of Vision-Language Models Under Adversarial Frequency-Domain Perturbations}
\author{\IEEEauthorblockN{Jordan Vice}

\and
\IEEEauthorblockN{Naveed Akhtar}
\and
\IEEEauthorblockN{Yansong Gao}
\and
\IEEEauthorblockN{Richard Hartley}
\and
\IEEEauthorblockN{Ajmal Mian}
\thanks{Jordan Vice (jordan.vice@uwa.edu.au), Yansong Gao (garrison.gao@uwa.edu.au) and Ajmal Mian (ajmal.mian@uwa.edu.au) are with University of Western Australia. Naveed Akhtar (naveed.akhtar1@unimelb.edu.au) is with University of Melbourne. Richard Hartley (Richard.Hartley@anu.edu.au) is with Australian National University.}%
}

\maketitle

\begin{abstract}
Vision-Language Models (VLMs) are increasingly used as perceptual modules for visual content reasoning, including through captioning and DeepFake detection. 
In this work, we expose a critical vulnerability of VLMs when exposed to subtle, structured perturbations in the frequency domain. Specifically, we highlight how these feature transformations undermine authenticity/DeepFake detection and automated image captioning tasks. 
We design targeted image transformations, operating in the frequency domain to systematically adjust VLM outputs when exposed to frequency-perturbed real and synthetic images.
We demonstrate that the perturbation injection method generalizes across five state-of-the-art VLMs which includes different-parameter Qwen2/2.5 and BLIP models. Experiments on ten real and generated image datasets reveal that VLM judgments are sensitive to frequency-based cues and may not wholly align with semantic content. 
Crucially, we show that visually-imperceptible spatial frequency transformations expose the fragility of VLMs deployed for automated image captioning and authenticity detection tasks.
Our findings under realistic, black-box constraints challenge the reliability of VLMs, underscoring the need for robust multimodal perception systems.

\end{abstract}


\section{Introduction}
The rise of vision-enabled models has opened up novel possibilities for innovation in creativity, automation and technological accessibility. However, alongside these advancements comes growing concerns over their reliability and trustworthiness \cite{Xie2024, Kuang2023, Zhang2025}. 
Vision-Language Models (VLMs) are popular for their multimodal reasoning, aiding tasks like VQA, image captioning, and zero-shot classification \cite{Zhang2024a}. However, they are often used for perceptual tasks like misinformation detection and forensic analysis \cite{Li2024, Verdoliva2020}, despite not being explicitly optimized for such tasks \cite{Zhang2024a}.
The general appeal of VLMs lies in their apparent human-like reasoning, making them attractive as general-purpose AI assistants. This appeal poses a risk as users without sufficient domain knowledge may overestimate VLM reliability and reasoning capabilities \cite{Zhang2025,Dai2023}.

Images can be decomposed into spatial frequency components, each reflecting different levels of visual structure \cite{Campbell1968, Valois1988, Chen2019, Li2023a}. Low frequencies capture coarse features like lighting gradients and overall shapes, while high frequencies encode fine details, textures, and edges \cite{Campbell1968, Valois1988, Chen2019}, often useful for distinguishing real from fake images \cite{Durall2020}. The mid-frequency band likely corresponds to object-level features, bridging background composition and textural detail. By leveraging these properties, we can introduce frequency-based perturbations to induce adversarial behavior in VLMs. Exposing such vulnerabilities is critical, particularly as VLMs are increasingly deployed in safety-sensitive contexts \cite{Ilyas2019, Zhang2024b}. In this work, we examine how frequency-domain perturbations compromise VLM reliability in two scenarios: (i) visual authenticity assessment via high-frequency manipulation, and (ii) image captioning via mid-frequency perturbations.

As generative models become integrated in mainstream applications, distinguishing real from synthetic content becomes crucial for media integrity and reliable decision-making in high-stakes settings \cite{Li2024, Verdoliva2020, Guo2025}.
Here, `authenticity detection' encompasses whether VLMs can reliably assess if an input image is real, functioning as a DeepFake detector. We demonstrate that their performance in this task is fundamentally unreliable. 
When evaluating high-fidelity images generated using state-of-the-art models (e.g., SD3.5), VLMs often misclassify synthetic content as real. 
We further expose their unreliability by applying structured frequency-domain noise that shifts the model’s predictions, causing synthetic images to be classified as “real” and real images as “generated.”

\begin{figure}
    \centering
    \includegraphics[width=0.95\linewidth]{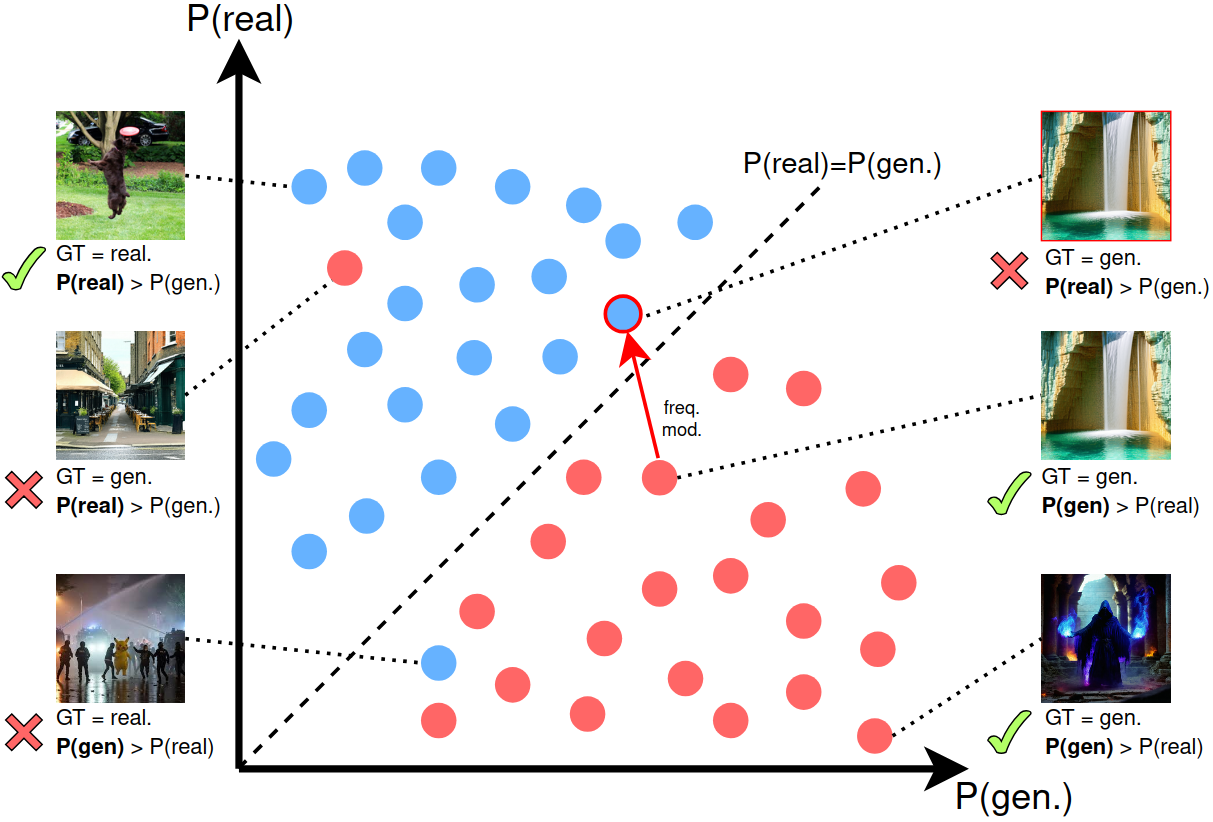}
    \caption{An abstract 2D space where images are positioned based on how likely a VLM perceives their realism. Blue points represent $\text{GT}$=real images, and red points represent $\text{GT}$=generated ones. The diagonal line separates predictions: images above it are judged more ``real" than ``generated'' and vice-versa. Example images show both correct and incorrect classifications. Applying spatial frequency perturbations can shift a generated image (red point) across this boundary, causing the model to misinterpret it as real. GT = ground truth.}
    \label{FIG_decision_boundary}
    \vspace{-4mm}
\end{figure}

Figure~\ref{FIG_decision_boundary} illustrates VLM interpretations of real vs. synthetic images, highlighting how high-fidelity generated samples are naturally classified as real.
Cases like the waterfall example in Fig. \ref{FIG_decision_boundary} are clearly synthetic to the human eye, but can then be pushed beyond the decision boundary through adversarial spatial frequency transformations.
Our findings suggest that the decision boundaries learned by VLMs for distinguishing real and generated content are surprisingly fragile, particularly under image quality-preserving perturbations. This raises the hypothesis that VLMs, especially those with fewer parameters, may rely on frequency-domain cues as a proxy for authenticity, rather than grounding their predictions in semantic content. For models that advertise natural reasoning capabilities, highlighting this key vulnerability becomes crucially important.

\begin{figure}
    \centering
    \includegraphics[width=0.95\linewidth]{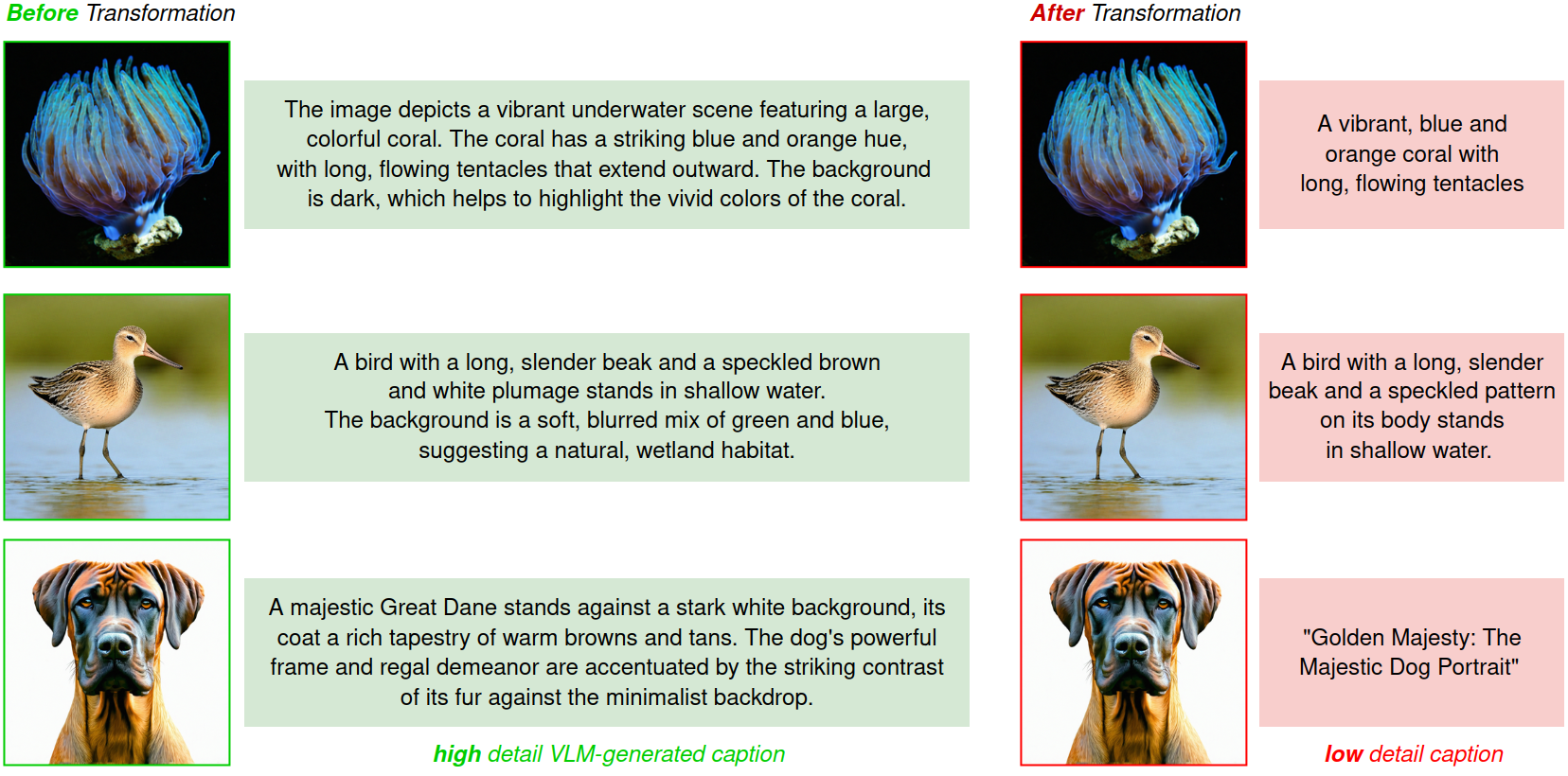}
    \caption{We illustrate how (imperceptible) mid-frequency perturbations manipulate the semantic richness of image captions generated by VLMs. Each row shows an input image and the VLM-generated caption before (left) and after (right) perturbations are applied. While original images elicit rich, context-aware descriptions, perturbations in the frequency domain can lead to shorter and less informative outputs.}
    \label{FIG_caption_motivation}
    \vspace{-4mm}
\end{figure}

We also leverage spatial frequency transformations to undermine the reliability of VLMs for image captioning tasks. We find that by manipulating mid-frequency components, we can effectively degrade the quality of generated image captions, as visualized in Fig. \ref{FIG_caption_motivation}. Naturally, the manipulation of VLM captions aligns with related resource exhaustion/latency attacks \cite{Chen2024, Pietrantuono2023}. By applying perturbations and monitoring the movement of VLM caption embeddings w.r.t. the original output (using CLIP \cite{Radford2021}), we can effectively control caption detail.
To investigate these vulnerabilities in a realistic scenario, we evaluate several VLMs under black-box constraints, wherein the attacker has access only to model inputs and outputs, i.e., input image, textual query and the textual VLM output. 
The application of spatial frequency perturbations operates entirely without access to model weights or gradients, making it compatible with the black-box attack constraints.

The imperceptible perturbations are designed to incrementally shift the model’s output toward a target realism score or CLIP embedding dissimilarity threshold.
By leveraging generative models for both image synthesis and vision-language reasoning, we suggest that future advances in both T2I and VLM architectures may give rise to a perpetual cat-and-mouse dynamic between generation and perception. We demonstrate how introducing frequency domain perturbations into this dynamic will give a competitive edge to the adversary.

To summarize our contributions: 
(\textit{i}) We highlight the unreliability of VLMs for visual authenticity (DeepFake) detection tasks, particularly when exposed to high-quality synthetic images generated by state-of-the-art models. To facilitate our investigations, we consider a continuous ``realism likelihood'' authenticity detection case. Manipulating high-frequency components exacerbates this confusion and is applicable for manipulating perceptions of both real and generated images.
(\textit{ii}) We extend our frequency perturbation framework to automated image captioning, showing that mid-frequency domain perturbations can significantly degrade the semantic richness of VLM-generated captions without altering the image's high-level perceptual content. To validate the choice of each perturbation range, we conduct cross-task transferability ablations which confirm our choice of ranges.
(\textit{iii}) We show that sparse, structured perturbations in the frequency domain can reliably manipulate model predictions, revealing a systemic vulnerability in how VLMs interpret visual inputs. Our findings suggest that their reasoning is fundamentally tied to low-level image structure rather than high-level semantics.
(\textit{iv}) We release the RGFreq Dataset \cite{RGFreqDataset}, comprised of real, generated and frequency-perturbed images to support future work in strengthening the robustness of multimodal models when performing authenticity detection and VLM-enabled image captioning tasks.

\section{Background and Related Works}
\noindent\textbf{Text-to-Image Generation Models} build on 
architectures like Generative Adversarial Networks (GANs) \cite{Goodfellow2014} and Variational Autoencoders (VAEs) \cite{Kingma2014}.
Diffusion models improved generation quality, leveraging pre-defined noise processes to incrementally denoise samples from a known, Gaussian distribution to target image representations \cite{Ho2020}. 
Unconditional models like Denoising Diffusion Probabilistic Model (DDPM) \cite{Ho2020} and DDIM \cite{Song2021}, now serve as a generative backbone for conditional models. 
Multimodal conditioning networks facilitate finer control over generation in diffusion models. Proprietary text-to-image models like DALL-E \cite{Ramesh2022} and Imagen \cite{Saharia2022} pair token-based transformers (e.g., CLIP \cite{Radford2021}, T5 \cite{Raffel2020}) with large conditional diffusion models, typically comprising one or more U-Nets \cite{Ronneberger2015}.
Stable Diffusion \cite{Rombach2022} operates in a learned latent space, separating the computational burden of diffusion from the high-dimensional pixel space. 
Successive improvements have produced variants like V1 \cite{Rombach2022}, XL \cite{Podell2023}, and V3 \cite{Esser2024}. 

When guided by structured scene descriptions, these models can produce photorealistic and semantically coherent images that may be indistinguishable from real photographs to both humans and vision-enabled intelligent systems. This capability raises important questions about the trustworthiness of visual content, especially in downstream applications that rely on perceptual systems for classification or authenticity detection. 

\noindent\textbf{Vision-Language Models} typically comprise a vision encoder - based on a convolutional neural network (CNN) \cite{He2016} or Vision Transformer (ViT) \cite{Dosovitskiy2021}, paired with a language model or multimodal fusion module. Trained on large-scale image–text corpora, they align visual and linguistic representations, which underpinned models designed for tasks like Visual Question Answering (VQA) \cite{Antol2015} and image captioning.
Generalization and zero-shot capabilities are then improved through CLIP \cite{Radford2021}, which introduces contrastive learning between image/text pairs at scale. Joint representations learned by CLIP demonstrates how vast internet data can be leveraged to better interpret complex visual concepts. This has  been extended and refined in models like BLIP \cite{Li2023} which enables open-ended multimodal reasoning through complex semantic spaces.

Through their general-purpose design, VLMs have increasingly been deployed as perceptual systems within broader foundation model ecosystems \cite{Bommasani2021}, including content moderation and visual retrieval \cite{Kiela2020,Lee2018}.
Models like LLaVA \cite{Liu2023}, Qwen-VL \cite{Bai2023, Wang2024}, and Gemma \cite{Beyer2024} have shown that aligning frozen vision encoders with LLMs through lightweight adapters or instruction tuning enables conversational VLMs.
Their deployment for instruction-based scene understanding tasks has positioned VLMs as powerful (and popular) tools. Here, we look to identify VLM reliability concerns and define the limits of their image captioning and authenticity detection capabilities when subjected to imperceptible perturbations.

\noindent\textbf{Synthetic Content Detection}.
While VLMs have advanced significantly, they are increasingly applied beyond their original design scope, including in tasks like visual content authenticity detection. 
This requires nuanced understanding of real vs. generated data distributions and resilience to minor perturbations.
Detecting synthetic content is a critical and evolving research area. Early approaches leveraged generator-specific artifacts and frequency-domain discrepancies between real and fake images \cite{Zhou2018, Durall2020}. Benchmarks such as FaceForensics++ \cite{Rossler2019} and the Facebook DeepFake Detection Challenge dataset \cite{Dolhansky2020} have driven progress. Notably,  \cite{Durall2020} demonstrated that upsampling operations in GANs introduce spectral distortions, enabling CNN-based detectors to exploit frequency cues.

Nevertheless, diffusion models now generate photorealistic images that evade such detection techniques \cite{Coccomini2024, Wang2023, Chandra2025}, as GAN-specific artifacts often do not generalize \cite{Coccomini2024}. To address this, Wang et al. \cite{Wang2023} proposed DIffusion Reconstruction Error (DIRE), showing that synthetic images are reconstructed more precisely than real ones, leading to improved generalization across model types. Similarly, SynthBuster \cite{Bammey2024} analyzes high-frequency spectral inconsistencies in diffusion-generated content for robust detection. While frequency-based features are central to many vision tasks, the susceptibility of VLMs to spatial-domain perturbations, despite their growing role in realness detection, remains underexplored.
Tariq et al.~\cite{Tariq2025} evaluated VLMs in a zero-shot setting for deepfake detection across face-swap, reenactment, and synthetic image categories. We extend this line of inquiry by introducing adversarially-guided spatial frequency perturbations and assessing their impact across two distinct tasks. As synthetic imagery increasingly resembles real content, reliable detection remains an open challenge. Our findings expose fundamental vulnerabilities in VLMs under black-box conditions, revealing unstable semantic reasoning when subjected to structured attacks.

\begin{figure*}
    \centering
    \includegraphics[width=0.95\linewidth]{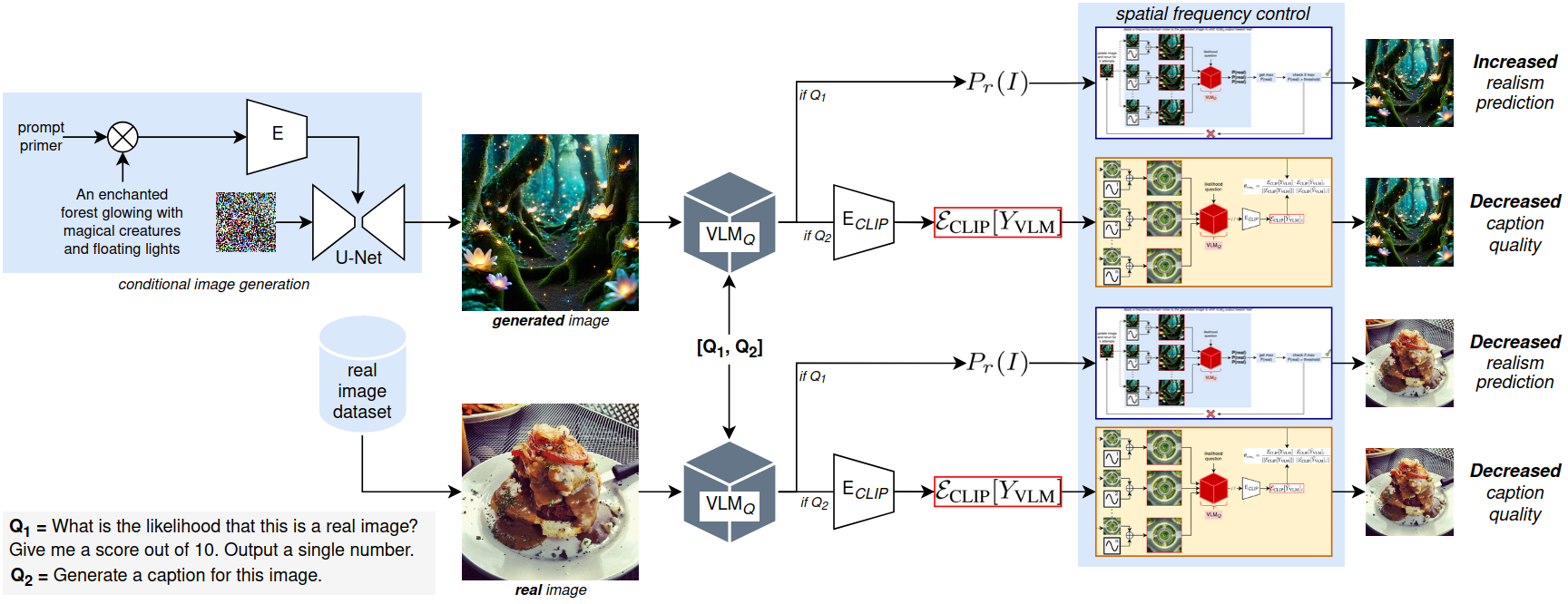}
    \caption{We explore how VLM outputs are influenced by changes in frequency domain features across two tasks: (i) authenticity detection and (ii) automated image captioning. Querying VLMs with unperturbed images (synthetic or real) may result in $P_r(I)$ prediction in-line with the ground truth (e.g. $P_r(I)=\frac{4}{10}$). However, subtle spatial frequency perturbations can induce unreliable VLM behavior, shifting the VLM output across the decision boundary ($P_r(I)=\frac{4\rightarrow8}{10}$). Likewise, perturbations are also applied to manipulate the quality of VLM-generated captions. We expand on the specifics for each task in Figs. \ref{FIG_perturbed_realism} and \ref{FIG_perturbed_caption}.}
    \label{FIG_high_level}
    \vspace{-4mm}
\end{figure*}

\noindent\textbf{Adversarial Vulnerabilities in Visual Models}.
Szegedy et al. \cite{Szegedy2014} first demonstrated that even high-accuracy image classifiers can be manipulated by small, imperceptible input changes. Follow-up work attributed such vulnerabilities to the linear nature of deep networks, leading to gradient-based attacks like the Fast Gradient Sign Method (FGSM) \cite{Goodfellow2015} and iterative attacks like Projected Gradient Descent (PGD) \cite{Madry2018}. 

Vision-language models are equally susceptible.
Li et al. \cite{Li2021} showed that human-generated, adversarially phrased questions can significantly reduce VQA accuracy. Similarly, adding imperceptible noise to the background of an image can alter VQA answers \cite{Chaturvedi2020}. Sophisticated VLMs also appear to be vulnerable to multimodal adversarial attacks \cite{Zhang2025, Guo2025}. In a black-box setting, adversarial image-text pairs crafted against a CLIP or BLIP model can transfer to fool systems like BLIP-2 or MiniGPT-4, producing targeted incorrect responses \cite{Zhao2023}. Moreover, by exploiting the alignment between image and text embeddings, an attack can jointly perturb an image and adjust a text prompt to remain semantically consistent while misleading the model’s alignment objective \cite{Ye2024}.


Zhang et al.\cite{Zhang2025} highlighted the fragility of VLMs under perceptible image perturbations and adversarial textual inputs. While their focus lies on pronounced perturbations, our work emphasizes imperceptibility, which is a critical factor for real-world applicability. Constraining attacks to high-frequency components is know to enhance stealth while preserving efficacy\cite{Luo2022}. Frequency-aware methods have also improved adversarial transferability across models~\cite{Yang2024}, reinforcing the significance of spectral cues. Additionally, diffusion models have been leveraged to generate visually plausible adversarial examples~\cite{Dai2024, Guo2025, Kang2025}, and latent-space manipulations (e.g., facial attribute editing) have yielded natural-looking yet misleading outputs~\cite{Hu2024}. These findings underscore the pervasive adversarial vulnerability in vision-enabled systems.

\section{Methodology}
We introduce a high-level summary of our decision-guided, black-box approach in Fig. \ref{FIG_high_level}, outlining how frequency components can be exploited to manipulate VLM decisions. We expand on how sparse frequency perturbations are applied for manipulating perceptions of authenticity in Fig. \ref{FIG_perturbed_realism}, adopting a similar approach to expose the vulnerabilities in image captioning applications as well. 
We first define key concepts related to T2I models and VLMs. Then, we elaborate on the design of our frequency-based black-box perturbation method.

\subsection{Preliminaries}
\begin{figure}
    \centering
    \includegraphics[width=0.95\linewidth]{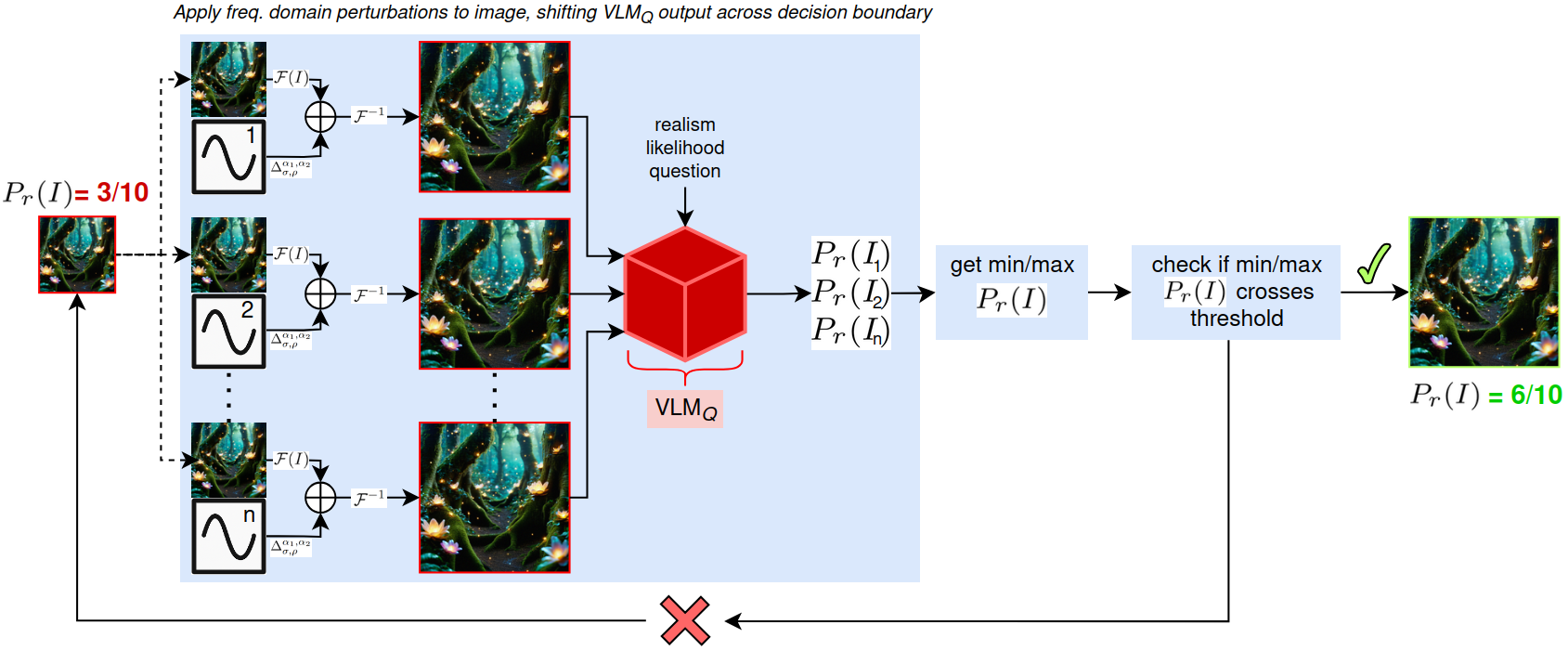}
    \caption{For authenticity detection, we apply sparse, \textit{high}-frequency perturbations to an image in an iterative loop, querying VLM for its realness prediction at each step. Multiple candidate perturbations are evaluated, and the highest $P_r(\Tilde{I})$ is selected per iteration, dependent on the target. Here, we demonstrate how exploiting the sensitivity of VLMs to frequency-domain cues enables subtle manipulation of synthetic content toward a “real” classification.}
    \label{FIG_perturbed_realism}
    \vspace{-4mm}
\end{figure}

\begin{figure*}
    \centering
    \includegraphics[width=0.95\linewidth]{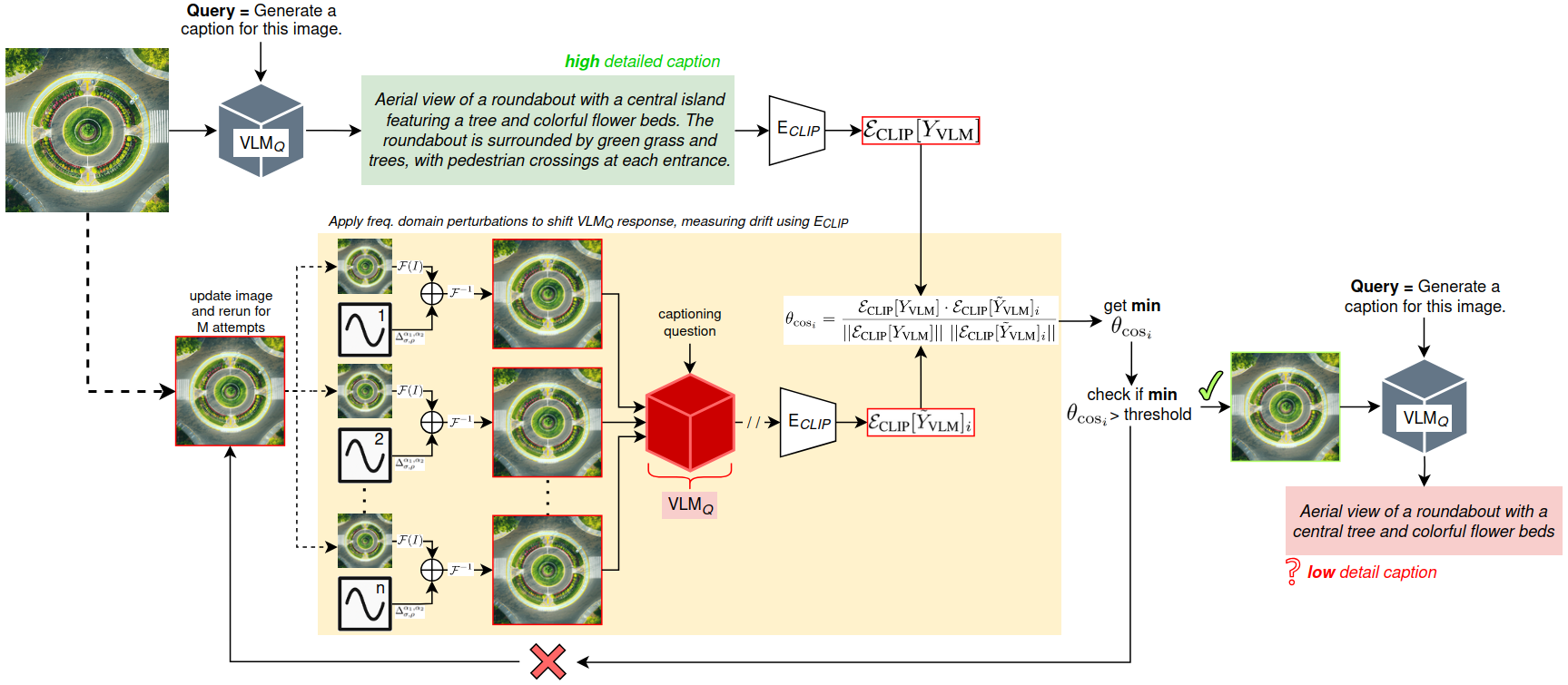}
    \caption{We assess the robustness of VLM-generated captions under mid-frequency perturbations by minimizing the CLIP similarity score `$\theta_{\cos}$' between original and perturbed caption embeddings. At each iteration `$i$', candidate perturbations are constructed in the frequency domain, and the perturbed image yielding the caption with the lowest $\theta_{\cos}$ is selected. The process repeats until $\theta_{\cos}$ reaches its goal - or the max number of iterations is reached. At which point, the optimally-perturbed image is chosen. This procedure reveals that imperceptible changes in the mid-frequency region can manipulate VLM behavior.}
    \label{FIG_perturbed_caption}
    \vspace{-4mm}
\end{figure*}

\noindent\textbf{Definitions.}
As shown in Fig. \ref{FIG_high_level}, our method is guided by VLM decisions. We introduce a plug-and-play spatial frequency control block to enable iterative, structured transformations in the frequency domain\, operating in a black-box manner. We expand on our perturbation methods in Figs. \ref{FIG_perturbed_realism} and  \ref{FIG_perturbed_caption}.

We use state-of-the-art image generation models to create a new, synthetic image dataset.
Let $x_p$ and $x_q$ denote the (textual) conditional image generation prompt and VLM input query, respectively. Our approach operates within the context of images, hence;  let $I_R$ and $I_G$ denote real and generated images, respectively, i.e., $I_R,I_G \in \mathbb{R}^{H \times W \times c=3}$. The VLM leverages $x_q$ and $I_{R/G}$ inputs to generate an output response $Y_{\text{VLM}}$. The task is to assess the reliability of VLM outputs when a small spatial frequency perturbations $f(\cdot)$ are applied to an image, keeping $x_q$ constant. We posit that minor spatial frequency transformations will undermine VLM reliability.

\noindent\textbf{Text-to-Image Models} allow for the generation of realistic, semantically-aligned images, leveraging inputs prompts $x_p$ and rich latent and text embedding spaces. Incorporating a text encoder `$\mathcal{E}_x(\cdot)$' (often CLIP \cite{Ramesh2022}) allows for the projection of a tokenized input prompt $x_p$ onto a learned embedding space. Through iterative denoising, the conditional diffusion model $\mathcal{M}_{\text{D}}(\cdot)$ generates an $x_p$-aligned image representation from an initial noise sample $\mathcal{N}_0$ across $T_\text{D}$ time-steps. This process is supplemented by a guidance scale which we denote by `$\gamma$', which controls the strength of conditioning. We can functionally represent the `$I_G$' image generation process as
\begin{equation}
    I_G = \mathcal{M}_{\text{D}}(\mathcal{E}_x[x_p],\gamma, \mathcal{N}_t,t) ~\forall~ t \in T_\text{D}.
\end{equation}

\noindent\textbf{Vision-Language Models} are $[x_q,I_{R/G}]$-input models, outputting a textual response `$Y_{\text{VLM}}$'. Encoders `$\mathcal{E}_x(\cdot)$' and `$\mathcal{E}_v(\cdot)$' process text and image inputs, respectively, projecting their respective inputs onto learned feature spaces $\in \mathbb{R}^{d}$. Cross-modal alignment in VLM training processes optimize the positioning of encoded features on high-dimensional manifolds \cite{Radford2021,Li2022}. A multimodal reasoning model `$\mathcal{M}_{\text{VL}}$' combines encoded features to output a response,
\begin{equation}
    Y_{\text{VLM}} = \mathcal{M}_{\text{VL}}(\mathcal{E}_x[x_q],\mathcal{E}_v[I_{R/G}]).
\end{equation}
This generalized formulation describes both contrastive models (like CLIP), which uses cosine similarity in a shared embedding space for zero-shot prediction \cite{Radford2021} and fusion-based models such as BLIP-2, which employ cross-attention to integrate visual and textual modalities in instruction-tuned reasoning settings \cite{Li2023}. In this work, we focus on the latter, which outputs a natural language response.

\subsection{Spatial Frequency Perturbations}
Prior works have shown that CNNs are biased toward high-frequency textures and patterns rather than global semantic structures \cite{Geirhos2019,Ilyas2019}, making them susceptible to imperceptible perturbations that preserve pixel-space appearance but alter frequency-domain information. High-frequency changes are often unnoticeable to human eyes. This legislates the use of the frequency domain for adversarial attacks and may highlight the reliance of machine vision systems on frequency-domain indicators. Durall et al. discussed how the frequency spectra of generated images differ from natural distributions \cite{Durall2020}. Logically, vision models could therefore learn and inherently rely on frequency heuristics, due to training data features.

We explore how VLM responses change when exposed to imperceptible changes in the spatial frequency domain. 
As visualized in Figs. \ref{FIG_perturbed_realism} and \ref{FIG_perturbed_caption}, VLM outputs guide the optimization of applied frequency perturbations. For authenticity detection, we bind the VLM output to a 10-point scale\footnote{where `$Y_{\text{VLM}}=0$', is a confidently \textbf{!real} prediction and `$Y_{\text{VLM}}=10$', denotes a confidently \textbf{real} prediction} and divide scores into three bins: (\textit{i}) $Y_{\text{VLM}}<\tau_1$, (\textit{ii}) $\tau_1 <Y_{\text{VLM}}<\tau_2$ and, (\textit{iii}) $Y_{\text{VLM}} > \tau_2$, where $\{\tau_1,\tau_2\}$=\{4, 6\}. Responses in the $\{\tau_1\rightarrow\tau_2\}$ range suggest some confusion due to images residing close to the real/!real cluster boundary. As discussed, for a fixed query $x_q$, we expect that frequency perturbations will disrupt $Y_{\text{VLM}}$, shifting the likelihood distribution across the three bins. We denote the \textit{adversarial} response as $\Tilde{Y}_{\text{VLM}}$.

To undermine the reliability of automated image captioning, the goal is to spatially perturb the image such that the VLM-generated caption changes (see Figs. \ref{FIG_caption_motivation}, \ref{FIG_perturbed_caption}). We employ an auxiliary CLIP encoder `$\mathcal{E}_{CLIP}(\cdot)$' to project the original and adversarial VLM outputs $\{Y_{\text{VLM}},\Tilde{Y}_{\text{VLM}}\}$ onto a shared CLIP space. We compare the original and adversarial caption embeddings and apply a similarity threshold `$\tau_{\text{sim}}=0.5$' which controls the amount of spatial frequency perturbations required to change the VLM caption. Our perturbations here are guided by the drift in response \textit{w.r.t.} the original, measured through cosine \textit{dis-}similarity between the CLIP embeddings.

To formalize our spatial frequency perturbations as a transformation on an image, let $I_{R/G}\in\mathbb{R}^{H\times W \times3}$ be the input image and $\psi_{\sigma,\rho,\alpha_1,\alpha_2}(I_{R/G})$ be the frequency-domain perturbation operator, where:
`$\sigma$' = standard deviation of injected frequency noise,
`$\rho$' = sparsity ratio controlling number of perturbed points and,
`$\alpha_1,\alpha_2 \in [0,1]$' = normalized lower and upper bounds on radial frequency magnitude.
Thus, we define the frequency-perturbed image $\Tilde{I}_{R/G}$ as:
\begin{equation}
    \Tilde{I}_{R/G} = \mathcal{F}^{-1}(\mathcal{F}(I_{R/G})+\Delta^{\alpha_1,\alpha_2}_{\sigma,~\rho}),
\end{equation}
where $\mathcal{F}(\cdot)$ and $\mathcal{F}^{-1}(\cdot)$ represent forward and inverse Fourier transforms, respectively. $\Delta^{\alpha_1,\alpha_2}_{\sigma,\rho}$ is the sparse noise matrix applied in the $\alpha_1\rightarrow\alpha_2$ frequency band. 
Spatial analysis works typically equate high frequencies to sharpness/detail and low frequencies to coarse image features/background \cite{Campbell1968, Shapley1985,Vuilleumier2003}. Typical spatial frequency curves are quadratic with steep slopes and narrow mid-frequency ranges \cite{Campbell1968, Shapley1985}. These seminal works inform our frequency range design.
We define the \textit{high} frequency range as points within \{$\alpha_1$=0.85, $\alpha_2$=1.00\}. For mid-frequency, we define a narrower \{$\alpha_1$=0.49, $\alpha_2$=0.51\}.
Given the general higher concentration of low-frequency image features, we found that low-frequency transformations result in perceptible changes in the image which harms our imperceptibility requirement. Thus, we only consider high- and mid-frequency ranges when applying perturbations. Parameters `$\sigma$, $\rho$' change w.r.t. input shape, where;
\begin{itemize}
    \item $\sigma=0.025\times H\times W $, i.e., 2.5\% standard deviation, proportional to input size;
    \item $\rho=0.1\times H\times W $, i.e., 10\% data points transformed, also proportional to input size.
\end{itemize}

Given a target VLM response/threshold $Y_{\tau}$, we construct a perturbation $\Delta^{\alpha_1,\alpha_2}_{\sigma,~\rho}$ and corresponding perturbed sample $\Tilde{I}_{R/G}$ to induce an adversarial response $\Tilde{Y}_{\text{VLM}}$ that approaches, or deviates from $Y_{\tau}$, depending on the task. 

\begin{figure*}
    \centering
    \includegraphics[width=0.9\linewidth]{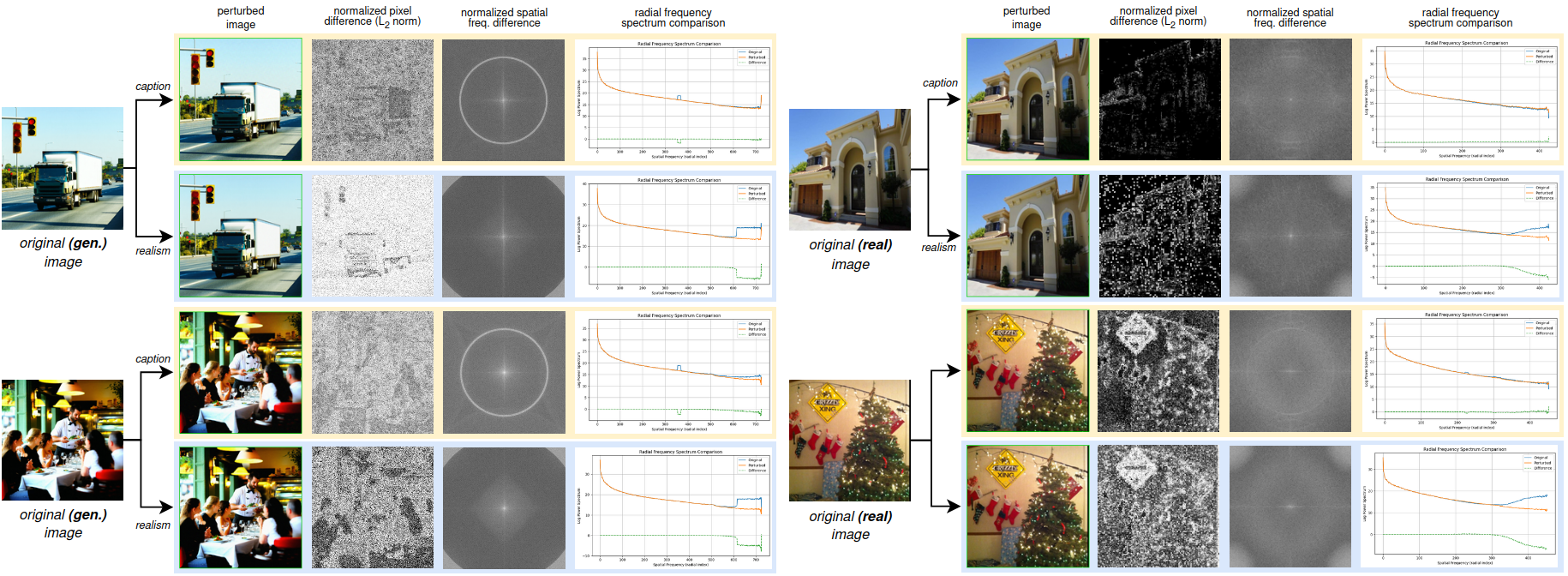}
    \caption{Visual comparison of perturbation effects on generated and real images for assessing caption- and realism-based reliability. Each row shows: (1) the perturbed image, (2) the binary, $L_2$-normalized pixel-wise difference, (3) the spatial frequency domain difference - which visualizes $\alpha_1,\alpha_2$ frequency bounds and, (4) the corresponding radial frequency spectrum comparison of original vs. perturbed images. We see that perturbations on generated images yield structured differences in both pixel and frequency domains, while perturbations on real images often produce more localized and higher-sparsity changes.}
    \label{FIG_qualitative_results}
    \vspace{-4mm}
\end{figure*}

Since we operate in a black-box setting, the perturbation is optimized through iterative querying and selection over candidate perturbations drawn from a constrained band-limited frequency space. At each iteration, we sample a batch of candidate perturbations from the \{$\alpha_1$,$\alpha_2$\} frequency band. We apply each perturbation to the Fourier-transformed image and query the VLM, aiming to optimize th perturbation.
So, at each step $t$, we construct $N$ candidate perturbations, picking the \textit{best} candidate `$\Delta_t$' that moves the VLM output toward satisfying the criterion defined by $Y_{\tau}$ such that
\begin{equation}
    \Delta_t = \underset{\Delta_i \in \{\underset{1\rightarrow N}{\Delta}\}}{\arg\max}\mathcal{G}_\tau(\mathcal{M}_{VL}(\mathcal{E}_x[x_q],\mathcal{E}_v[\mathcal{F}^{-1}(\mathcal{F}(I)+\Delta_i)])
    ,Y_{\tau}),
\end{equation}
where $\Delta_i \in \underset{1\rightarrow N}{\Delta}= \{\Delta_1,...,\Delta_N\}$ defines the $N$ candidate perturbations. `$\mathcal{G}_\tau(\Tilde{Y}_{\text{VLM}},Y_{\tau})$' defines the goal function used to guide the adversarial response based on a target VLM output. The perturbation that best satisfies the goal `$\Delta_t$' is selected to update the image, iterating this process for $T_\Delta$ iterations, or until the goal has been satisfied. Recalling (3):
\begin{equation}
    \Tilde{I}_{t+1} = \mathcal{F}^{-1}(\mathcal{F}(\Tilde{I}_t+\Delta_t))~ \forall~ t \in T_\Delta-1.
\end{equation}
So, given a target VLM output/boundary value $Y_{\tau}$, we optimize `$\Delta^{\alpha_1,\alpha_2}_{\sigma,\rho}$' such that the adversarial decision $\Tilde{Y}_{\text{VLM}}$ approaches $Y_{\tau}$. We visualized this in Figs. \ref{FIG_perturbed_realism} and \ref{FIG_perturbed_caption}, highlighting its applicability in a black-box setup for two independent tasks.

\begin{figure}
    \centering
    \includegraphics[width=0.95\linewidth]{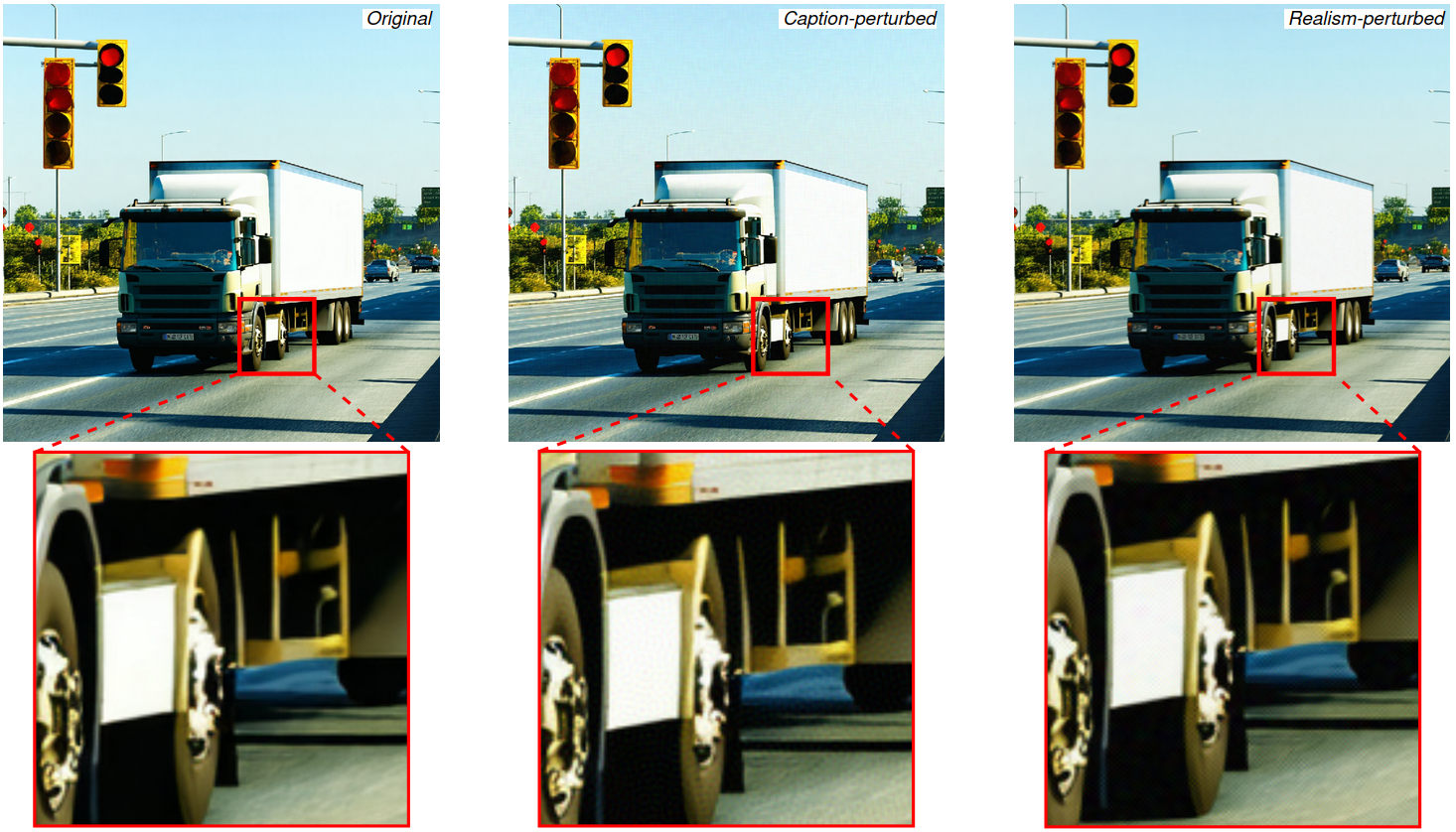}
    \caption{Localized image regions of \textbf{(left)} the original, \textbf{(middle)} captioning task and, \textbf{(right)} realism likelihood task outputs at 600\% magnification. Despite transformations applied in the spatial frequency domain, zoomed-in insets show that the perturbations remain largely imperceptible to the human eye.}
    \label{FIG_zoomed_inset}
    \vspace{-4mm}
\end{figure}

We illustrate the effects of spatial transformations in Fig. \ref{FIG_qualitative_results}, showing $L_2$ pixel differences and changes in spatial frequency between original vs. perturbed images. Affected frequency bands are radial, with clearer patterns appearing in \textit{generated} images. Although binary difference maps highlight altered pixels, perturbations are imperceptible, even under significant magnification (see Fig.\ref{FIG_zoomed_inset}).
Interestingly, in Fig. \ref{FIG_qualitative_results} we see \textit{softer} spatial frequency differences in real images, which reflect their statistical regularities. In contrast, generated images provide a more fragile manifold and contain repeated textures/features, allowing perturbations to have more visible fingerprints in the frequency domain.  Bammey et al. \cite{Bammey2024} exploited frequency features to detect diffusion-generated content. Compounded with our findings, this suggests that frequency cues play a critical role in discerning real from synthetic content and the overall image understanding in VLMs.

\noindent\textbf{Goal Function Definition.} For authenticity detection, the goal is to shift VLM realness prediction across a binary threshold, effectively flipping the predicted class. We define the goal function $\mathcal{G}_\tau(\cdot)$ based on the ground truth output $Y_{\text{GT}}$ and the direction in which the output should be moved. Specifically:
\begin{align}
\mathcal{G}_\tau(\Tilde{Y}_{\text{VLM}},Y_{\tau}) = 
\begin{cases}
    \tau_{2}+1 ~ \text{if}~ Y_\text{GT} \leq 5, \\
    \tau_{1}-1 ~ \text{if}~ Y_\text{GT} > 5,
\end{cases}
\end{align}
where $\tau_1$ and $\tau_2$ are scalar targets used to push the VLM output toward the opposing class. In practice, we use $\tau_1$=4 and $\tau_2$=6 to span the full output scale. The perturbation is optimized to drive $\Tilde{Y}_{\text{VLM}}$ toward $\mathcal{G}_\tau$, thereby inverting the VLM's decision.

To propagate unreliability in captioning, the task is to shift the VLM-generated caption away $Y_{\text{GT}}$, minimizing the cosine similarity of CLIP embeddings. To achieve this, we deploy the following for each candidate perturbation
\begin{equation}
\mathcal{G}_\tau(\Tilde{Y}_{\text{VLM}},Y_{\tau}) = \theta_{\cos_i}=\frac{\mathcal{E}_{\text{CLIP}}[Y_{\text{GT}}] \cdot \mathcal{E}_{\text{CLIP}}[\Tilde{Y}_{\text{VLM}}]_i}{ ||\mathcal{E}_{\text{CLIP}}[Y_{\text{GT}}]||~||\mathcal{E}_{\text{CLIP}}[\Tilde{Y}_{\text{VLM}}]_i||}.
\end{equation}
We continue to minimize similarity until spatial frequency-perturbed image satisfies `$\theta_{\cos_i}<\tau_{\text{sim}}$' i.e., a $\tau_{\text{sim}}=50\%$ dissimilarity \textit{w.r.t.} the original caption\footnote{or until the maximum number of iterations has been reached.}. We capture $\theta_{\cos_i}$ and the number of tokens in $\Tilde{Y}_{\text{VLM}}$ to measure (i) the impact of manipulating the mid-spatial frequency range and, (ii) the semantic detail in the updated caption.

\section{Experimental Setup}
\noindent\textbf{High Fidelity Image Generation.} We deploy Stable Diffusion v3.5-Large (SD3.5) \cite{stablediffusion35} model to generate high fidelity as part of our experiments. We identify three scene classes which facilitate prompt construction in diverse settings, namely; (i) fantasy scenes, e.g., ``A sword embedded in a stone surrounded by enchanted mist'', (ii) realistic outdoor scenes, e.g., ``A self-driving car making a left turn at an urban intersection'' and, (iii) ImageNet class-labeled scenes, e.g., ``affenpinscher, monkey pinscher, monkey dog''. We utilize GPT-4o to construct prompts for fantasy and realistic outdoor scenes. The full collection of scene prompts is available \href{https://github.com/JJ-Vice/Adversarial-Spatial-Perturbations-For-VLMs}{on GitHub}.
As visualized previously in Fig. \ref{FIG_high_level}, we apply a \textit{prompt primer} to better guide the model toward `realistic' representations. The full prompt becomes: ``\textit{Photo realistic image of  \textit{\{SCENE\}}. Include details and clarity, Perspective, Realistic Colors and Contrast. No Visible Artifacts or Filters. Contextual Recognition.}''. We generate $N$ unique images per prompt, adjusting the random seed, generating images over $T_\text{D}$=100 steps and using a constant guidance scale of $\gamma$=9.0 (recalling (1)).

\begin{figure}
    \centering
    \includegraphics[width=0.95\linewidth]{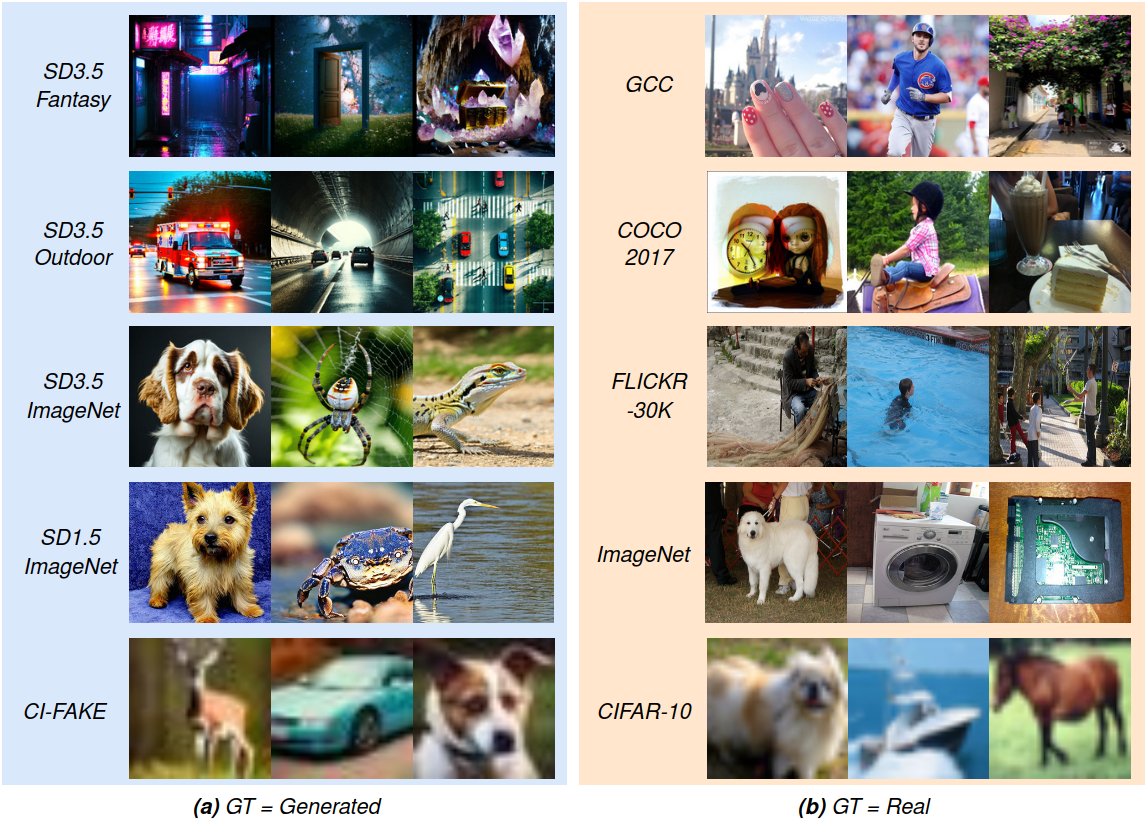}
    \caption{Sample images from all of the \textbf{(a)} generated datasets and \textbf{(b)} real image datasets evaluated in this work.}
    \label{FIG_dataset_images}
    \vspace{-4mm}
\end{figure}

\noindent\textbf{Datasets.}
In addition to the SD3.5 generated image dataset described above, our evaluations also consider Stable ImageNet-1K \cite{Kinakh2022}, which includes ImageNet class-labeled scenes generated using the older SD1.4 model \cite{Rombach2022}. Similarly, the CIFAKE dataset \cite{Bird2023} adopts a similar approach, generating a synthetic version of the popular CIFAR-10 dataset \cite{Krizhevsky2009} (which we also evaluate). Both image realism and caption generation tasks can be completed irrespective of whether the image is generated or not. Thus, evaluations using real-world datasets are also necessary. To that end, we leverage popular image  datasets, including: (i) Google Conceptual Captions (GCC) \cite{Sharma2018} (ii) the 2017 Microsoft Common Objects in Context (COCO-2017) \cite{Lin2014}, (iii) Flickr30k \cite{Plummer2015}, (iv) \textit{real} ImageNet \cite{Russakovsky2015} and as discussed, (v) CIFAR-10 \cite{Krizhevsky2009}. Having real and synthetic versions of CIFAR-10 and ImageNet datasets offers an intuitive comparison across the two tasks.
Logically, image realism likelihood experiments greatly benefit from having real and synthetic (ground truth) counterparts. Fig. \ref{FIG_dataset_images} provides representative image samples from each dataset.

\noindent\textbf{Implementation Details.} For our primary experiments, we use the Qwen2-VL-7B-Instruct VLM \cite{Wang2024} as our query/victim model. When finding optimal spatial frequency perturbations, the number of candidate perturbations changes depending on the task. The introduction of the CLIP encoder in the image captioning task increases computational load and thus, we extract ten unique candidate perturbations per step. For the image realism task (which only requires the query VLM), we generate twenty candidate perturbations per step. For both tasks, we apply perturbations for a maximum of five steps \textit{or} until the $\mathcal{G}_\tau(\Tilde{Y}_{\text{VLM}},Y_{\tau})$ goal is reached. Increasing the number of candidate perturbations and steps widens the search range for optimal perturbations in a black-box setting, at the cost of increased inference time. 
We deploy conservative values to account for the breadth of our experiments and ablations.

\begin{table*}
    \centering
    \resizebox{0.9\linewidth}{!}{%
    \begin{tabular}{l|lccc|ccc|ccc|cc}
         Dataset & Type & $\overline{P_r(I)}$ & $\overline{P_r(\Tilde{I})}$ & $\Delta \overline{P_r}$ & $P_r(I) < \tau_1$ & $\tau_1 \leq P_r(I) \leq \tau_2 $ & $P_r(I) > \tau_2$ & $P_r(\Tilde{I}) < \tau_1 $ & $\tau_1 \leq P_r(\Tilde{I}) \leq \tau_2 $ & $P_r(\Tilde{I}) > \tau_2 $  & $P_r(I)$=real & $P_r(\Tilde{I})$=real \\
         \hline
         & & \multicolumn{11}{c}{Qwen2-VL-7B-Instruct} \\
         \hline
         SD3.5-Fantasy           & Gen.      & \underline{2.4283} \tiny{$\pm 1.8359$} & \underline{2.9667} \tiny{$\pm \mathbf{1.9821}$} & $\uparrow$ 0.5384 & \textbf{0.8100} & 0.1817 & 0.0083 & \textbf{0.7100} & 0.2683 & 0.0217 & 0.1450 & 0.2050 \\
         SD3.5-Outdoor           & Gen.      & 6.5409 \tiny{$\pm 1.2089$} & 7.2091 \tiny{$\pm 1.1316$} & $\uparrow$ \textbf{0.6682} & 0.0261 & 0.6659 & 0.3080 & 0.0080 & 0.3932 & 0.5989 & \textbf{0.9591} & \textbf{0.9920} \\
         SD3.5-ImageNet          & Gen.      & 7.4727 \tiny{$\pm 1.9181$} & 7.8412 \tiny{$\pm 1.5611$} & $\uparrow$ 0.3685 & 0.0740 & 0.2315 & 0.6945 & 0.0424 & 0.1454 & 0.8121 & 0.9152 & 0.9539 \\
         \textit{SD1.5}-ImageNet & Gen.      & 7.8569 \tiny{$\pm \mathbf{2.1465}$} & 8.0655 \tiny{$\pm 1.9457$} & $\uparrow$ 0.2086 & 0.0827 & 0.1464 & 0.7709 & 0.0692 & \underline{0.1135} & 0.8173 & 0.8649 & 0.9169 \\
         CIFAKE                  & Gen.      & 5.6921 \tiny{$\pm \underline{1.0070}$} & 5.7796 \tiny{$\pm \underline{0.9792}$} & $\uparrow$ 0.0875 & 0.0704 & \textbf{0.9296} & \underline{0.0000} & 0.0598 & \textbf{0.9280} & \underline{0.0123} & 0.8943 & 0.9141 \\
         GCC                     & Real      & 8.4370 \tiny{$\pm 1.4512$} & 8.0686 \tiny{$\pm 1.6183$} & $\downarrow$ 0.3684 & 0.0000 & 0.1758 & 0.8242 & 0.0119 & 0.2220 & 0.7661 & 0.9139 & 0.8957\\
         COCO-2017               & Real      & \textbf{9.1807} \tiny{$\pm 1.4224$} & \textbf{8.4831} \tiny{$\pm 1.3317$} & $\downarrow$ 0.6976 & 0.0000 & 0.0886 & 0.9114 & 0.0035 & 0.1208 & \textbf{0.8757} & 0.9159 & 0.9104 \\
         FLICKR-30k              & Real      & 9.1279 \tiny{$\pm 1.3645$} & 8.4039 \tiny{$\pm 1.2955$} & $\downarrow$ \underline{0.7240} & 0.0000 & \underline{0.0875} & \textbf{0.9125} & \underline{0.0010} & 0.1421 & 0.8569 & 0.9226 & 0.9176 \\
         CIFAR10                 & Real      & 6.2772 \tiny{$\pm 1.0868$} & 5.6034 \tiny{$\pm 1.4589$} & $\downarrow$ 0.6738 & 0.0000 & 0.8680 & 0.1320 & 0.1029 & 0.8355 & 0.0616 & 0.8900 & 0.7151 \\
         ImageNet                & Real      & 8.7467 \tiny{$\pm 1.5602$} & 8.1711 \tiny{$\pm 1.5176$} & $\downarrow$ 0.5756 & 0.0000 & 0.1507 & 0.8493 & 0.0058 & 0.2030 & 0.7912 & 0.8985 & 0.8891 \\ 
         \hline
    \end{tabular}}
    \caption{Comparison of Realness likelihood for different generated image test cases. $\overline{P(r)}$ denotes the average realness likelihood for images in a test set. We also report the proportion of samples that report VLM-realness likelihoods: (i) below $\tau_1$, (ii) between $\tau_1$ and $\tau_2$ and, (iii) beyond  $\tau_2$, where $\tau_1=3$ and $\tau_2=6$. We also present binary realism results here as well.
    Model = \textbf{Qwen2-VL-7B-Instruct}. \underline{underline} = lowest, \textbf{bold} = highest .}
    \label{TABLE_realness}
    \vspace{-4mm}
\end{table*}
\noindent\textbf{Metrics.} We aim to assess reliability across two VLM tasks: perceiving image realism and automated image captioning. 

To evaluate how VLMs perceive real vs. synthetic content, we first report `$\overline{P_r(I)}$' and `$\overline{P_r(\Tilde{I})}$' which denote the mean realism scores of real and generated test sets, respectively. Then, we report how realism scores are distributed across `$Y_{\text{VLM}}<\tau_1$', `$\tau_1 <Y_{\text{VLM}}<\tau_2$' and `$Y_{\text{VLM}} > \tau_2$' bins. This evaluation helps in identifying how confident the model is in its prediction, where \textit{real} image sets should have little-to-no samples in the $Y_{\text{VLM}}<\tau_1$. In comparison, if there are a large number of generated samples that reside in the $Y_{\text{VLM}} > \tau_2$ range, this indicates that the VLM is unreliable when identifying synthetic content. By applying spatial-frequency perturbations, we demonstrate that the distributions can be manipulated. Finally, we report binary realism scores, where $Y_{\text{VLM}}>5\Rightarrow P_r(I)\text{=real}$ and, $Y_{\text{VLM}}<5\Rightarrow P_r(I)\text{=!real}$

To evaluate image captioning performance, we capture the length of the VLM-generated caption and the semantic drift \textit{w.r.t.} to the original caption before and after applying spatial frequency perturbations, recalling (7). Reporting the length of the caption allows us to determine how verbose $Y_{\text{VLM}}$ is when exposed to manipulated spatial frequencies, with real-world relevance as it pertains to resource exhaustion attacks \cite{Chen2024, Pietrantuono2023}. 
By modeling the mean drift $\overline{\theta_{\cos}}$, we validate our hypothesis that mid-frequency regions, situated at the intersection of object boundaries and background, can undermine the reliability of scene understanding tasks like image captioning.
Semantic drift also measures the susceptibility of the test set to transformations, which may depend on the dataset distribution.




\section{Results}
Here, we present our reliability evaluations, supporting the hypothesis that spatial frequency perturbations in high- and mid-frequency bands can compromise VLM reliability across two distinct tasks. We further examine task transferability, e.g., whether realism-optimized \textit{mid}-frequency perturbations similarly impact image captioning, and whether \textit{high}-frequency perturbations influence VLM-perceived realism. To validate the generality of our approach beyond the primary model, we assess its effectiveness across multiple VLMs, analyzing reliability as a function of model size and architecture.

\subsection{Image Authenticity Assessments}
As evidenced in the frequency domain representations in Fig. \ref{FIG_qualitative_results}, real and generated images present interesting frequency domain properties, especially when subjected to guided perturbations. While imperceptible to the naked eye (see Fig. \ref{FIG_zoomed_inset}), perturbing images in high-frequency bands does have a demonstrable effect on the perception of image authenticity/realism in VLMs. As discussed, we achieve this by shifting the VLM output \textit{w.r.t.} the ground truth label, i.e., from real$\rightarrow$ !real and vice-versa. Using a ten-point scale and guidance terms $\tau_1=4$ and $\tau_2=6$, we show that high spatial frequency features can be exploited to flip VLM predictions, without compromising the visual integrity of the image. 

We report our primary findings on the Qwen2-VL-7B-Instruct model in Table \ref{TABLE_realness}, where
we see that our guided frequency perturbations consistently manipulate VLM perceptions of realism and effectively shift samples across decision boundaries without patching the image with visible perturbations.
As per $P_r(I)$ range observations (bounded by $\{\tau_1,\tau_2\}$), the VLM will reliably output high scores $>\tau_2$ for real image sets. This represents a \textit{highly confident} prediction of perceived realism. However, as per $P_r(\Tilde{I})$ results, the applied perturbations add uncertainty to predictions. We observe that initially, VLMs can struggle to discern real images from high-fidelity generated content. Adding high frequency-bounded perturbations amplifies this, adding greater uncertainty to authenticity perception. So, for the SD3.5 Fantasy subset, which contains clearly-fictitious generated scenes (see Fig. \ref{FIG_dataset_images}), the 6\% increase in binary realism predictions and distribution of scores presents \textit{significant} unreliability. 

\begin{figure*}
    \centering
    \includegraphics[width=0.95\linewidth]{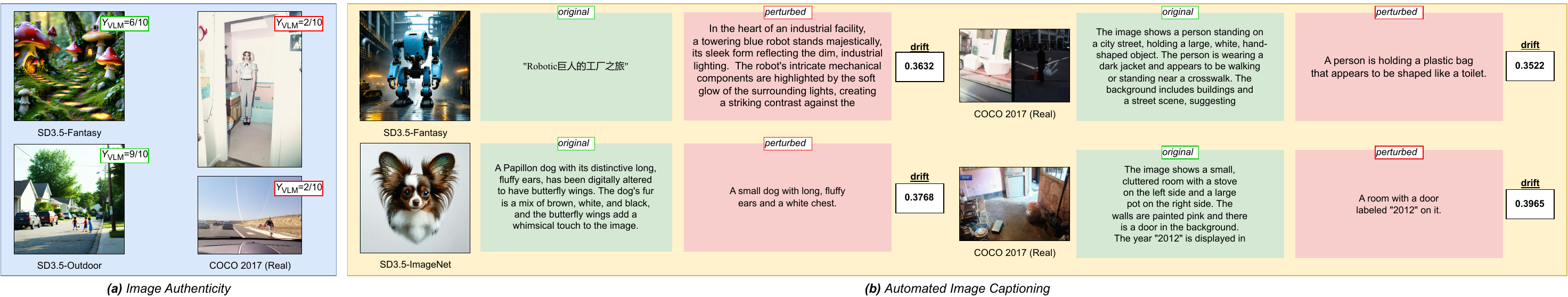}
    \caption{Visualizing representative outcomes across both tasks. \textbf{(a)} Real and generated test images alongside their perceived realness scores after applying high-spatial frequency perturbation. \textbf{(b)} Sample test images (real and generated) with corresponding VLM-generated captions pre- and post-perturbation.}
    \label{FIG_interesting_findings}
    \vspace{-4mm}
\end{figure*}

The $\Delta \overline{P_r}$ column formalizes the changes in realism likelihoods across test cases. We see that VLMs are more sensitive to perturbations applied to \textit{real} images, despite the visual imperceptibility. This demonstrates that VLMs do not reliably assess semantic content, but may be influenced by underlying image characteristics instead. As previously discussed, high-frequency components typically encode fine-grained details, textures and edges - all of which are commonplace in real image distributions. Therefore, by actively targeting the high-spatial frequency regions to undermine VLM image realism evaluations, it is logical that real images may be highly sensitive. Quantitative results reported on additional VLMs later support this claim. Across generated image sets, the SD3.5-generated outdoor scenes present a 0.6628 increase in VLM-likelihood score, which demonstrates how close these images may be to the real/!real decision boundary. This hypothesis is supported by the $P_r(\Tilde{I})$=real column in Table \ref{TABLE_realness}, where only 0.8\% samples classify as synthetic (higher than all GT=real image sets), despite the dataset being entirely generated.

Comparing real and synthetic image pairs is crucial. For the real vs. !real ImageNet sets, high-frequency perturbations reduce the maximum difference in $\overline{P_r(I)}$ likelihood from 1.2740 to 0.3299. Similarly, for CI-FAKE vs. CIFAR-10, the gap narrows from 0.5851 to 0.1762. This highlights the effectiveness of our approach. By applying high-frequency perturbations, we augment perceptions of real and synthetic images, leading to significant confusion in VLMs, which undermines their reliability for authenticity detection.


We identify interesting cases in Fig. \ref{FIG_interesting_findings}, showing perturbed images and their realism scores. Notably, SD3.5-Fantasy images can be perceived as “real” after perturbation, while actual photos from the COCO dataset are misclassified as “not real” despite appearing realistic to the human eye. 
While VLMs can be unreliable authenticity detectors naturally, these capabilities can be undermined through applied perturbations in high spatial frequency components.
We show that a decision-guidance term ($\{\tau_1,\tau_2\}$) can be used to control VLM realism perceptions, without damaging the image. This exposes a technical shortcoming in how VLMs assess authenticity, i.e., statistical features have a strong influence over decision-making, more-so than the actual semantic content. Our reported results validate our intuition that high-frequency features are important for realism assessment tasks. We explore this further when conducting ablation studies on cross-sample transferability.

\subsection{Automated Image Captioning}
Automated captioning tasks have been improved through VLM-integration, which has accelerated training processes. However, we pose a question on how reliable VLMs are when images are exposed to \textit{mid}-spatial frequency perturbations. 

Similar to image realism experiments, we evaluate our approach across generated and real image datasets. We report our results in Table \ref{TABLE_caption}.
Perturbing frequency-domain components enables the manipulation of VLM captioning without introducing perceptible artifacts. We evaluate the impact of our method by analyzing changes in caption verbosity, using $\Delta_{\text{length}}(Y_{\text{VLM}})$ and $\Delta_{N_\text{tokens}}(Y_{\text{VLM}})$ to capture variations in string length and token count, respectively.
Then, we consider the impact that perturbations have on VLM semantic scene understanding, using semantic drift. Here, we measure (cosine) dissimilarity between original `$Y_{\text{VLM}}$' and perturbed `$\Tilde{Y}_{\text{VLM}}$' outputs. 

\begin{table}
    \centering
    \resizebox{0.9\linewidth}{!}{%
    \begin{tabular}{l|lcccc}
         Dataset & Type & $\Delta_{\text{length}}(Y_{\text{VLM}})$ & $\overline{N_\text{tokens}}$ & $\Delta_{N_\text{tokens}}(Y_{\text{VLM}})$ & $Y_{\text{VLM}}$ drift \\
         \hline
         & \multicolumn{4}{c}{Qwen2-VL-7B-Instruct} \\
         \hline
         SD3.5-Fantasy          & Gen.      & \textbf{6.1467}\tiny{$\pm \mathbf{93.6376}$} & 23.673 \tiny{$\pm 11.220$} & \textbf{1.2083}\tiny{$\pm 15.3846$}     & 0.2537\tiny{$\pm 0.1543$} \\
         SD3.5-Outdoor          & Gen.      & -8.9272\tiny{$\pm 78.4638$}                  & 19.634 \tiny{$\pm 9.831$}  & -1.1409\tiny{$\pm 13.6620$}             & 0.2386\tiny{$\pm 0.1267$} \\
         SD3.5-ImageNet         & Gen.      & -20.8485\tiny{$\pm 81.7229$}                 & 22.773 \tiny{$\pm 9.376$}  & -2.8861\tiny{$\pm 13.8570$}             & 0.2358\tiny{$\pm 0.1290$} \\
         \textit{SD1.5}-ImageNet& Gen.      & \underline{-44.1040}\tiny{$\pm 66.6996$}     & 31.745 \tiny{$\pm 10.949$} & \underline{-8.1061}\tiny{$\pm 11.9185$} & 0.1447\tiny{$\pm 0.1247$}\\
         CIFAKE                 & Gen.      & 0.8685\tiny{$\pm \underline{31.1339}$}       & 16.166 \tiny{$\pm 11.559$} & 0.1310\tiny{$\pm \underline{5.5173}$}   & \underline{0.0502}\tiny{$\pm \underline{0.1053}$} \\
         GCC                    & Real      & -11.9594\tiny{$\pm 58.8067$}                 & 31.369 \tiny{$\pm 11.843$} & -1.9509\tiny{$\pm 10.2663$}             & 0.1072\tiny{$\pm 0.1106$} \\
         COCO-2017              & Real      & -34.6565\tiny{$\pm 70.4028$}                 & 35.265 \tiny{$\pm 11.830$} & -6.3925\tiny{$\pm 12.6866$}             & 0.1451\tiny{$\pm 0.1339$} \\
         FLICKR-30k             & Real      & -34.3595\tiny{$\pm 76.2959$}                 & 34.465 \tiny{$\pm 12.764$} & -6.3040\tiny{$\pm 13.6887$}             & 0.1764\tiny{$\pm 0.1492$} \\
         CIFAR10                & Real      & 5.6420\tiny{$\pm 90.6681$}                   & 21.169 \tiny{$\pm 13.163$} & 0.9345\tiny{$\pm \mathbf{15.9941}$}     & \textbf{0.2596}\tiny{$\pm \mathbf{0.1852}$} \\
         ImageNet               & Real      & -28.2024\tiny{$\pm 72.6559$}                 & 35.318 \tiny{$\pm 11.085$} & -5.2214\tiny{$\pm 12.9174$}             & 0.1355\tiny{$\pm 0.1284$} \\ 
         \hline
    \end{tabular}}
    \caption{Comparison of VLM caption verbosity and semantic drift. We report the overall string length and number of tokens in original and perturbed cases, capturing this change via $\{\Delta_{\text{length}},\Delta_{N_\text{tokens}}\}$. Mean semantic drift of CLIP embeddings is calculated as $1-\overline{\theta_{\cos}}$. Lower $\{\Delta_{\text{length}},\Delta_{N_\text{tokens}}\}$ indicate less detailed captions. 
    $\uparrow\text{drift}$ = larger perturbation influence.}
    \label{TABLE_caption}
    \vspace{-4mm}
\end{table}

\begin{table*}[]
    \centering
    \resizebox{0.9\linewidth}{!}{%
    \begin{tabular}{llcc|cccc|cc}
          $\mathcal{F}(\cdot)$ & original task & ($\alpha_1,\alpha_2$) & Type & $\overline{P_r(I)}$  & $P_r(I) < \tau_1$ & $\tau_1 \leq P_r(I) \leq \tau_2 $ & $P_r(I) > \tau_2$ & $P_r(I)=\text{real}$ &$Y_{\text{VLM}}$ drift\\
          \hline
          \multirow{2}{*}{None} & \multirow{2}{*}{base image} &  \multirow{2}{*}{($0.00,0.00$)}  & Real & 8.3832\tiny{$\pm 1.7577$} & 0.0000 & 0.2703 & 0.7297 & 0.9082 \\
                                                               &  & & Gen. & 6.5852\tiny{$\pm 2.3373$} & 0.1343 & 0.4381 & 0.4276 &  0.8284 & \\
        \hline
          \multirow{2}{*}{Mid} & \multirow{2}{*}{captioning} & \multirow{2}{*}{($0.49,0.51$)}    & Real & 8.0088\tiny{$\pm 1.5946$} & 0.0111 & 0.2740 & 0.7150 & 0.9354 & 0.1678\tiny{$\pm 0.1542$} \\
                                                             &    & & Gen. & 6.6080\tiny{$\pm 2.3438$} & 0.1378 & 0.4286 & 0.4337 & 0.8372 & 0.1496\tiny{$\pm 0.1441$} \\
        \hline
          \multirow{2}{*}{High} & \multirow{2}{*}{realism} & \multirow{2}{*}{($0.85,1.00$)}   & Real & 7.7618\tiny{$\pm 1.8020$} & 0.0246 & 0.3009 & 0.6745 & 0.8663 & 0.1326\tiny{$\pm 0.1350$} \\
                                                            &     & & Gen. & 6.8672\tiny{$\pm 2.2109$} & 0.1113 & 0.3872 & 0.5015 & 0.8683 & 0.0912\tiny{$\pm 0.1230$} \\
         \hline
    \end{tabular}}
    \caption{Cross-task Transferability, assessing changes in realism and image captioning capabilities under different spatial frequency perturbation conditions, recalling that $\alpha_1,\alpha_2$ denote the upper and lower frequency band used to apply the perturbation.}
    \label{TABLE_transferability}
    \vspace{-4mm}
\end{table*}
While differentiating real vs. not real is not pivotal in \textit{this} task, VLMs are more likely to have been trained on a larger proportion of real data. This imbalance may lead to learned priors that influence captioning behavior under perturbation.
As highlighted in Table \ref{TABLE_caption}, perturbing the hyper-\textit{unrealistic} SD3.5-Fantasy set results in increases in the verbosity of the output, while also boasting the second highest $Y_{\text{VLM}}$ drift value. This underscores the uniqueness of the samples represented in this set in comparison to others. 
The likelihood of fantastical scenes appearing in the training set of the VLM is not as high as other image sets and thus, adding perturbations appears to have pushed the VLM to the closest outputs that were semantically different but higher in verbosity. 
We highlight an example of this in Fig. \ref{FIG_interesting_findings} where for the generated image of the robot, we see that (i) the original $Y_{\text{VLM}}$ contains Chinese text which translate to ``Journey Through the Giant’s Factory'' and (ii) the perturbed equivalent produces a significantly more detailed caption.

We note the minimal impact on the CIFAKE dataset as reported in Table \ref{TABLE_caption}, evidencing only 5\% drift and a minimal change in verbosity. 
We found that the length of VLM-generated outputs for CIFAKE and CIFAR-10 were lowest in comparison to others. This observation is likely related to the size of the images (32$\times$32 pixels). Such a small number of features limits the describable content. Furthermore, the deployed VLM would not be optimized for handling images of such a small size. As a result, when applying spatial frequency perturbations, the search space for optimally-perturbed images is reduced. Despite less features, our method is still effective, particularly as CIFAR-10 reports the highest $Y_{\text{VLM}}$ drift.

The guiding parameter could have been any of four options presented in Table \ref{TABLE_caption}, depending on the intended effect of the perturbations. We selected semantic drift as it (i) reflects caption detail, and (ii) tracks how well the VLM captures image semantics, which should be independent of low-level statistics or frequency perturbations. In practice, any observable metric could serve this purpose.
Picking any of the length-based metrics would make the method applicable for a resource exhaustion attack \cite{Chen2024, Pietrantuono2023} (or mitigation) setup.
We previously hypothesized that manipulating mid-spatial frequency components could influence how VLMs interpret object–background interactions. Even narrow perturbations (2\% band, $\alpha_1$=0.49,$\alpha_2$=0.51) significantly impact captioning while leaving visual features largely intact (see Figs. \ref{FIG_qualitative_results}, \ref{FIG_zoomed_inset}, \ref{FIG_interesting_findings}). This highlights the sensitivity of VLMs to the spatial frequency components of learned images. 

\subsection{Ablation Study: Cross-task Transferability}
Here, our aim is to check: (i) if \textit{mid}-frequency perturbations cause major changes to realness likelihoods, despite being optimized for manipulating captioning performance and, (ii) if \textit{high}-frequency perturbations have any effect on how the VLM captions the image. We report the results in Table \ref{TABLE_transferability}, merging real and generated image sets into single categories.
High-frequency perturbations shift VLM perceptions of realism in expected inverse directions, causing real images to output `!real' and vice-versa.
When suboptimal mid-frequency perturbations are applied, the aggregated $P_r(I)$ likelihood distributions governed by $\tau_1$ and $\tau_2$ remain largely unchanged. It is worth noting that for real images, despite $\overline{P_r(I)}$ reducing (from None$\rightarrow$Mid), the $P_r(I)=\text{real}$ increases. This points to the shape of the VLM output distribution and a concentration around the boundary condition i.e. $P_r(I)\equiv\tau_1$. This is supported by the reduction in standard deviation from $\pm 1.758$ to $\pm 1.595$, which evidences a narrower distribution curve. 

For detecting realism/authenticity of generated images, we see that there is minimal change in observations from base to mid-frequency results of $\Delta \overline{P_r(I)}=0.023$ and almost no change in how VLM-generated $P_r(I)$ scores are distributed. As intended, these changes are much more prevalent when high-spatial frequency perturbations are applied. This supports our initial hypothesis that high frequency components are more critical for VLM perceptions of realism. This is due to the inherent structure of real-world data, whereas recurring structural components are more prevalent in generated images.

In assessing VLM captioning capabilities under mid- and high-spatial frequency perturbations, we find that the former is more effective. The larger mean $Y_{\text{VLM}}$ drift and slightly greater standard deviations across real and generated sets suggests that the targeted perturbations in mid-spatial frequency components have a wider-ranging impact on VLM perceptions of image content. High-frequency perturbations while still having some effect, result in more similar VLM-generated captions, particularly in generated images (lower $Y_{\text{VLM}}$ drift).
While we consider automated image captioning and image realism as two separate tasks with independent guidance terms, the extensions on this work are vast and could involve guidance across multiple VLM tasks. Optimizing spatial frequency perturbations across multiple tasks presents a logical next step and could lead to the design of universally-unreliable outputs that can fool VLMs across multiple tasks.

\subsection{Ablation Study: VLM Parameter Size vs. Reliability}
To assess the generalization ability of our  approach, we deploy similar authenticity detection and captioning evaluations on four additional VLMs, which vary in parameter size and architecture. Specifically, we evaluate (i) 2B-parameter variant of Qwen2-VL model \cite{Wang2024}, (ii) 3B-parameter Qwen2.5-VL model \cite{qwen2.5VL} and (iii/iv) 2.7B- and 6.7B-parameter BLIP2-VL models \cite{Li2022}.
We detail the results in Tables \ref{TABLE_captioning_ablation} and \ref{TABLE_realness_ablation}, which affirm that our method does generalize well and that model size can affect perturbation severity.

\begin{figure}
    \centering
    \includegraphics[width=\linewidth]{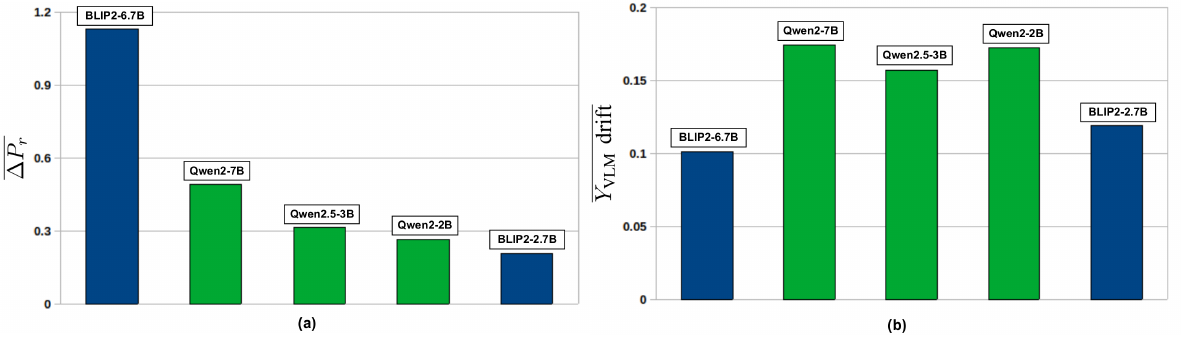}
    \caption{Average changes in VLM behavior when spatial perturbations are applied for: \textbf{(a)} perceiving image authenticity, modeled via mean $\Delta \overline{P_r}$ and, \textbf{(b)} automated image captioning, modeled via the mean $Y_{\text{VLM}}$ drift.}
    \label{FIG_ablation_bars}
    \vspace{-4mm}
\end{figure}

We quantify these observations through $\Delta \overline{P_r}$ and $Y_{\text{VLM}}$ drift columns in Tables \ref{TABLE_captioning_ablation} and \ref{TABLE_realness_ablation}. We identify interesting trends through reported mean characteristics of $\Delta \overline{P_r}$ and $Y_{\text{VLM}}$ drift metrics across models in Fig. \ref{FIG_ablation_bars}. Here, we observe that model parameter size has a large influence on the effectiveness of applying perturbations to adjust image authenticity assessments (see Fig. \ref{FIG_ablation_bars}(a)). In particular, the BLIP-6.7B model reports an absolute change of $\approx1.1$, whereas the 2.7B-variant reports an average change of $\approx0.2$. 
Logically, this can be explained through the granularity of feature mappings in small vs. lage parameter models and the influence this has on output diversity. 
We suspect that in larger-parameter VLMs, with richer learned feature spaces, the candidate perturbations have a higher chance of moving to a valid, point on the manifold, which influences the VLM output. In comparison, smaller-parameter models would logically have \textit{scattered} learned-feature spaces, which would require higher-strength perturbations to manipulate output representations. To maintain fair testing and our constraint that perturbations \textit{must be imperceptible}, our evaluations here use consistent perturbation hyper-parameters as defined previously. 

\begin{table}
    \centering
    \resizebox{0.9\linewidth}{!}{%
    \begin{tabular}{l|lcccc}
         Dataset & Type & $\Delta_{\text{length}}(Y_{\text{VLM}})$ & $\overline{N_\text{tokens}}$ &$\Delta_{N_\text{tokens}}(Y_{\text{VLM}})$ & $Y_{\text{VLM}}$ drift \\
         \hline
         & \multicolumn{4}{c}{Qwen2-VL-2B-Instruct} \\
         \hline
         SD3.5-Fantasy          & Gen.      & 0.791  \tiny{$\pm 38.694 $} & 12.584 \tiny{$\pm 4.911 $}  & 0.205  \tiny{$\pm 6.261 $} & 0.188 \tiny{$\pm 0.119 $}  \\
         SD3.5-Outdoor          & Gen.      & -0.888 \tiny{$\pm 32.274 $} & 12.421 \tiny{$\pm 4.060 $}  & 0.391  \tiny{$\pm 5.327 $} & 0.221 \tiny{$\pm 0.123 $}  \\
         SD3.5-ImageNet         & Gen.      & -1.700 \tiny{$\pm 32.992 $} & 10.150 \tiny{$\pm 4.209 $}  & -0.160 \tiny{$\pm 5.581 $} & 0.260 \tiny{$\pm 0.170 $}  \\
         \textit{SD1.5}-ImageNet& Gen.      & 2.455  \tiny{$\pm 23.425 $} & 11.428 \tiny{$\pm 4.003 $}  & -0.076 \tiny{$\pm 4.014 $} & 0.161 \tiny{$\pm 0.146 $}  \\
         CIFAKE                 & Gen.      & 0.610  \tiny{$\pm 13.024 $} & 9.200  \tiny{$\pm 3.140 $}  & 0.180  \tiny{$\pm 2.110 $} & 0.043 \tiny{$\pm 0.096 $}  \\
         GCC                    & Real      & 1.114  \tiny{$\pm 22.811 $} & 13.051 \tiny{$\pm 4.160 $}  & -0.025 \tiny{$\pm 4.163 $} & 0.119 \tiny{$\pm 0.126 $}  \\
         COCO-2017              & Real      & 8.960  \tiny{$\pm 26.327 $} & 12.210 \tiny{$\pm 4.942 $}  & 2.300  \tiny{$\pm 5.902 $} & 0.181 \tiny{$\pm 0.164 $}  \\
         FLICKR-30k             & Real      & 2.860  \tiny{$\pm 25.186 $} & 13.055 \tiny{$\pm 4.804 $}  & 0.850  \tiny{$\pm 5.634 $} & 0.187 \tiny{$\pm 0.151 $}  \\
         CIFAR10                & Real      & 2.990  \tiny{$\pm 26.363 $} & 9.725  \tiny{$\pm 3.500 $}  & 0.450  \tiny{$\pm 4.076 $} & 0.210 \tiny{$\pm 0.158 $}  \\
         ImageNet               & Real      & 7.561  \tiny{$\pm 26.952 $} & 12.199 \tiny{$\pm 4.629 $}  & 1.173  \tiny{$\pm 5.207 $} & 0.157 \tiny{$\pm 0.131 $}  \\
         \hline
         & \multicolumn{4}{c}{Qwen2.5-VL-3B-Instruct} \\
         \hline
         SD3.5-Fantasy          & Gen.      & 6.515  \tiny{$\pm 37.113 $} & 14.232 \tiny{$\pm 5.131 $}  & 1.152  \tiny{$\pm 6.348 $} & 0.200 \tiny{$\pm 0.139 $}  \\
         SD3.5-Outdoor          & Gen.      & -3.961 \tiny{$\pm 44.185 $} & 14.879 \tiny{$\pm 5.715 $}  & 0.475  \tiny{$\pm 7.690 $} & 0.204 \tiny{$\pm 0.122 $}  \\
         SD3.5-ImageNet         & Gen.      & -4.660 \tiny{$\pm 26.319 $} & 11.790 \tiny{$\pm 4.242 $}  & -0.740 \tiny{$\pm 4.306 $} & 0.186 \tiny{$\pm 0.111 $}  \\
         \textit{SD1.5}-ImageNet& Gen.      & 5.123  \tiny{$\pm 27.659 $} & 15.581 \tiny{$\pm 4.559 $}  & 1.292  \tiny{$\pm 4.622 $} & 0.136 \tiny{$\pm 0.123 $}  \\
         CIFAKE                 & Gen.      & 3.590  \tiny{$\pm 25.784 $} & 12.230 \tiny{$\pm 5.315 $}  & 0.520  \tiny{$\pm 4.665 $} & 0.147 \tiny{$\pm 0.183 $}  \\
         GCC                    & Real      & 2.101  \tiny{$\pm 29.950 $} & 16.006 \tiny{$\pm 5.495$}  & 0.367  \tiny{$\pm 5.321 $} & 0.109 \tiny{$\pm 0.100 $}  \\
         COCO-2017              & Real      & -2.050 \tiny{$\pm 26.269 $} & 14.980 \tiny{$\pm 5.450 $}  & -0.680 \tiny{$\pm 4.765 $} & 0.105 \tiny{$\pm 0.102 $}  \\
         FLICKR-30k             & Real      & -9.390 \tiny{$\pm 42.363 $} & 17.580 \tiny{$\pm 6.318 $}  & -1.820 \tiny{$\pm 7.603 $} & 0.150 \tiny{$\pm 0.125 $}  \\
         CIFAR10                & Real      & 2.120  \tiny{$\pm 28.052 $} & 11.815 \tiny{$\pm 4.659 $}  & 0.530 \tiny{$\pm 5.246 $}  & 0.214 \tiny{$\pm 0.191 $}  \\
         ImageNet               & Real      & 1.082  \tiny{$\pm 34.849 $} & 16.311 \tiny{$\pm 7.261 $}  & -0.378 \tiny{$\pm 5.953 $} & 0.123 \tiny{$\pm 0.096 $}  \\

         \hline
         & \multicolumn{4}{c}{Salesforce/BLIP2-2.7B} \\
         \hline
         SD3.5-Fantasy          & Gen.      & 0.862  \tiny{$\pm 32.409$} & 4.057 {$\pm 7.604$}  & 0.222  \tiny{$\pm 5.330 $}  & 0.059 \tiny{$\pm 0.161 $}  \\
         SD3.5-Outdoor          & Gen.      & 0.432  \tiny{$\pm 18.436$} & 5.682 {$\pm 5.439$}  & 0.063  \tiny{$\pm 3.504 $}  & 0.058 \tiny{$\pm 0.133 $}  \\
         SD3.5-ImageNet         & Gen.      & 2.150  \tiny{$\pm 42.401$} & 6.440 {$\pm 10.462$} & 0.500  \tiny{$\pm 7.839 $}  & 0.103 \tiny{$\pm 0.175 $}  \\
         \textit{SD1.5}-ImageNet& Gen.      & 8.679  \tiny{$\pm 59.685$} & 7.478 {$\pm 8.719$}  & 1.729  \tiny{$\pm 10.566 $} & 0.194 \tiny{$\pm 0.229 $}  \\
         CIFAKE                 & Gen.      & 0.770  \tiny{$\pm 9.420$}  & 8.340 {$\pm 8.385$}  & 0.080  \tiny{$\pm 1.548 $}  & 0.018 \tiny{$\pm 0.082 $}  \\
         GCC                    & Real      & 2.519  \tiny{$\pm 50.479$} & 9.278 {$\pm 10.855$} & 0.557  \tiny{$\pm 9.241 $}  & 0.090 \tiny{$\pm 0.176 $}  \\
         COCO-2017              & Real      & 5.770  \tiny{$\pm 47.483$} & 8.770 {$\pm 7.348$}  & 1.140  \tiny{$\pm 9.265 $}  & 0.156 \tiny{$\pm 0.196 $}  \\
         FLICKR-30k             & Real      & 2.470  \tiny{$\pm 41.880$} & 8.950 {$\pm 5.231$}  & 0.040  \tiny{$\pm 7.336 $}  & 0.239 \tiny{$\pm 0.173 $}  \\
         CIFAR10                & Real      & -2.680 \tiny{$\pm 34.360$} & 5.605 {$\pm 6.252$}  & -0.390 \tiny{$\pm 5.516 $}  & 0.093 \tiny{$\pm 0.162 $}  \\
         ImageNet               & Real      & 7.959  \tiny{$\pm 44.803$} & 6.372 {$\pm 5.790$}  & 1.316  \tiny{$\pm 7.345 $}  & 0.184 \tiny{$\pm 0.192 $}  \\
         \hline
         & \multicolumn{4}{c}{Salesforce/BLIP2-6.7B} \\
         \hline
         SD3.5-Fantasy          & Gen.      & -1.175 \tiny{$\pm 14.193$} & 3.143 \tiny{$\pm 3.118$}  & -0.226 \tiny{$\pm 2.398 $} & 0.040 \tiny{$\pm 0.110 $}  \\
         SD3.5-Outdoor          & Gen.      & 1.138  \tiny{$\pm 16.084$} & 5.434 \tiny{$\pm 5.150$}  & 0.267  \tiny{$\pm 3.312 $} & 0.059 \tiny{$\pm 0.119 $}  \\
         SD3.5-ImageNet         & Gen.      & -1.980 \tiny{$\pm 23.324$} & 3.460 \tiny{$\pm 4.907$}  & -0.500 \tiny{$\pm 5.865 $} & 0.054 \tiny{$\pm 0.115 $}  \\
         \textit{SD1.5}-ImageNet& Gen.      & 2.170  \tiny{$\pm 22.789$} & 3.264 \tiny{$\pm 2.980$}  & 0.231  \tiny{$\pm 3.562 $} & 0.101 \tiny{$\pm 0.144 $}  \\
         CIFAKE                 & Gen.      & 1.880  \tiny{$\pm 20.325$} & 6.580 \tiny{$\pm 7.168$}  & 0.380  \tiny{$\pm 3.776 $} & 0.014 \tiny{$\pm 0.069 $}  \\
         GCC                    & Real      & 4.380  \tiny{$\pm 29.305$} & 5.424 \tiny{$\pm 5.150$}  & 0.797  \tiny{$\pm 4.947 $} & 0.109 \tiny{$\pm 0.180 $}  \\
         COCO-2017              & Real      & -3.160 \tiny{$\pm 18.911$} & 6.300 \tiny{$\pm 6.187$}  & -0.420 \tiny{$\pm 3.919 $} & 0.171 \tiny{$\pm 0.154 $}  \\
         FLICKR-30k             & Real      & -3.500 \tiny{$\pm 23.895$} & 7.290 \tiny{$\pm 4.308$}  & -0.580 \tiny{$\pm 5.748 $} & 0.250 \tiny{$\pm 0.155 $}  \\
         CIFAR10                & Real      & -0.680 \tiny{$\pm 11.369$} & 4.520 \tiny{$\pm 3.074$}  & -0.120 \tiny{$\pm 2.078 $} & 0.064 \tiny{$\pm 0.111 $}  \\
         ImageNet               & Real      & -1.806 \tiny{$\pm 19.971$} & 4.801 \tiny{$\pm 3.544$}  & -0.194 \tiny{$\pm 3.626 $} & 0.154 \tiny{$\pm 0.144 $}  \\
         \hline
    \end{tabular}}
    \caption{Captioning ablation results across different parameter size models and architectures.}
    \label{TABLE_captioning_ablation}
    \vspace{-4mm}
\end{table}

\begin{table*}
    \centering
    \resizebox{0.9 \linewidth}{!}{%
    \begin{tabular}{l|lccc|ccc|ccc|cc}
         Dataset & Type & $\overline{P_r(I)}$ & $\overline{P_r(\Tilde{I})}$ & $\Delta \overline{P_r}$ & $P_r(I) < \tau_1$ & $\tau_1 \leq P_r(I) \leq \tau_2 $ & $P_r(I) > \tau_2$ & $P_r(\Tilde{I}) < \tau_1 $ & $\tau_1 \leq P_r(\Tilde{I}) \leq \tau_2 $ & $P_r(\Tilde{I}) > \tau_2 $  & $P_r(I)$=real & $P_r(\Tilde{I})$=real \\
         \hline
         & & \multicolumn{11}{c}{Qwen2-VL-2B-Instruct} \\
         \hline
         SD3.5-Fantasy           & Gen.      & 5.963 \tiny{$\pm 0.311$} & 6.007 \tiny{$\pm 0.116$} & $\uparrow$   0.044 & 0.010 & 0.990 & 0.000 & 0.000 & 0.997 & 0.003 & 1.000 & 1.000 \\
         SD3.5-Outdoor           & Gen.      & 6.211 \tiny{$\pm 0.615$} & 6.485 \tiny{$\pm 0.858$} & $\uparrow$   0.274 & 0.000 & 0.894 & 0.106 & 0.000 & 0.757 & 0.243 & 1.000 & 1.000 \\
         SD3.5-ImageNet          & Gen.      & 6.200 \tiny{$\pm 0.603$} & 6.420 \tiny{$\pm 0.819$} & $\uparrow$   0.220 & 0.000 & 0.900 & 0.100 & 0.000 & 0.790 & 0.210 & 1.000 & 1.000 \\
         \textit{SD1.5}-ImageNet & Gen.      & 8.213 \tiny{$\pm 1.415$} & 8.325 \tiny{$\pm 1.223$} & $\uparrow$   0.112 & 0.014 & 0.166 & 0.819 & 0.007 & 0.144 & 0.848 & 0.982 & 0.993 \\
         CIFAKE                  & Gen.      & 5.890 \tiny{$\pm 0.634$} & 5.890 \tiny{$\pm 0.634$} &              0.000 & 0.030 & 0.970 & 0.000 & 0.030 & 0.970 & 0.000 & 0.970 & 0.970 \\
         GCC                     & Real      & 6.987 \tiny{$\pm 1.115$} & 6.823 \tiny{$\pm 1.083$} & $\downarrow$ 0.165 & 0.000 & 0.519 & 0.481 & 0.000 & 0.595 & 0.405 & 0.975 & 0.975 \\
         COCO-2017               & Real      & 8.660 \tiny{$\pm 1.199$} & 8.010 \tiny{$\pm 1.352$} & $\downarrow$ 0.650 & 0.010 & 0.080 & 0.910 & 0.010 & 0.180 & 0.810 & 0.990 & 0.990 \\
         FLICKR-30k              & Real      & 8.810 \tiny{$\pm 0.720$} & 8.550 \tiny{$\pm 0.957$} & $\downarrow$ 0.260 & 0.000 & 0.060 & 0.940 & 0.000 & 0.110 & 0.890 & 1.000 & 1.000 \\
         CIFAR10                 & Real      & 5.890 \tiny{$\pm 0.584$} & 5.600 \tiny{$\pm 1.044$} & $\downarrow$ 0.290 & 0.030 & 0.970 & 0.000 & 0.130 & 0.870 & 0.000 & 0.960 & 0.870 \\
         ImageNet                & Real      & 8.347 \tiny{$\pm 1.293$} & 7.714 \tiny{$\pm 1.370$} & $\downarrow$ 0.632 & 0.010 & 0.143 & 0.847 & 0.010 & 0.276 & 0.714 & 0.990 & 0.990 \\ 
         \hline
         & & \multicolumn{11}{c}{Qwen2.5-VL-3B-Instruct} \\
         \hline
         SD3.5-Fantasy           & Gen.      & 2.230 \tiny{$\pm 0.836$} & 2.330 \tiny{$\pm 1.041$} & $\uparrow$   0.100 & 0.970 & 0.017 & 0.013 & 0.957 & 0.020 & 0.023 & 0.013 & 0.027 \\
         SD3.5-Outdoor           & Gen.      & 5.960 \tiny{$\pm 2.500$} & 6.739 \tiny{$\pm 2.149$} & $\uparrow$   0.779 & 0.344 & 0.098 & 0.558 & 0.204 & 0.089 & 0.706 & 0.603 & 0.736 \\
         SD3.5-ImageNet          & Gen.      & 4.820 \tiny{$\pm 2.564$} & 5.450 \tiny{$\pm 2.463$} & $\uparrow$   0.630 & 0.560 & 0.090 & 0.350 & 0.430 & 0.140 & 0.430 & 0.350 & 0.430 \\
         \textit{SD1.5}-ImageNet & Gen.      & 7.523 \tiny{$\pm 2.611$} & 7.827 \tiny{$\pm 2.385$} & $\uparrow$   0.303 & 0.188 & 0.014 & 0.798 & 0.155 & 0.007 & 0.838 & 0.798 & 0.838 \\
         CIFAKE                  & Gen.      & 4.290 \tiny{$\pm 1.684$} & 4.420 \tiny{$\pm 1.683$} & $\uparrow$   0.130 & 0.480 & 0.410 & 0.110 & 0.110 & 0.440 & 0.440 & 0.120 & 0.120 \\
         GCC                     & Real      & 6.544 \tiny{$\pm 2.881$} & 6.405 \tiny{$\pm 2.907$} & $\downarrow$ 0.139 & 0.278 & 0.051 & 0.671 & 0.291 & 0.051 & 0.658 & 0.696 & 0.684 \\
         COCO-2017               & Real      & 8.760 \tiny{$\pm 0.922$} & 8.410 \tiny{$\pm 1.436$} & $\downarrow$ 0.350 & 0.010 & 0.020 & 0.970 & 0.040 & 0.010 & 0.950 & 0.970 & 0.950 \\
         FLICKR-30k              & Real      & 8.240 \tiny{$\pm 1.415$} & 8.040 \tiny{$\pm 1.699$} & $\downarrow$ 0.200 & 0.040 & 0.010 & 0.950 & 0.060 & 0.000 & 0.940 & 0.950 & 0.940 \\
         CIFAR10                 & Real      & 4.340 \tiny{$\pm 1.805$} & 4.020 \tiny{$\pm 1.700$} & $\downarrow$ 0.320 & 0.510 & 0.350 & 0.140 & 0.580 & 0.320 & 0.100 & 0.150 & 0.100 \\
         ImageNet                & Real      & 8.378 \tiny{$\pm 1.447$} & 8.153 \tiny{$\pm 1.743$} & $\downarrow$ 0.224 & 0.041 & 0.020 & 0.939 & 0.071 & 0.010 & 0.918 & 0.939 & 0.918 \\ 
         \hline
         & &  \multicolumn{11}{c}{Salesforce/BLIP2-2.7B} \\
         \hline
         SD3.5-Fantasy           & Gen.      & 9.404 \tiny{$\pm 2.241$} & 9.669 \tiny{$\pm 1.697$} & $\uparrow$   0.265 & 0.066 & 0.000 & 0.934 & 0.037 & 0.000 & 0.963 & 0.934 & 0.963 \\
         SD3.5-Outdoor           & Gen.      & 9.186 \tiny{$\pm 2.584$} & 9.332 \tiny{$\pm 2.362$} & $\uparrow$   0.145 & 0.090 & 0.000 & 0.910 & 0.074 & 0.000 & 0.926 & 0.910 & 0.926 \\
         SD3.5-ImageNet          & Gen.      & 9.990 \tiny{$\pm 0.100$} & 9.990 \tiny{$\pm 0.100$} &              0.000 & 0.000 & 0.000 & 1.000 & 0.000 & 0.000 & 1.000 & 1.000 & 1.000 \\
         \textit{SD1.5}-ImageNet & Gen.      & 9.982 \tiny{$\pm 0.180$} & 9.993 \tiny{$\pm 0.121$} & $\uparrow$   0.011 & 0.000 & 0.000 & 1.000 & 0.000 & 0.000 & 1.000 & 1.000 & 1.000 \\
         CIFAKE                  & Gen.      & 8.218 \tiny{$\pm 3.559$} & 8.218 \tiny{$\pm 3.559$} &              0.000 & 0.192 & 0.000 & 0.808 & 0.192 & 0.000 & 0.808 & 0.808 & 0.808 \\
         GCC                     & Real      & 7.646 \tiny{$\pm 3.630$} & 7.519 \tiny{$\pm 3.633$} & $\downarrow$ 0.127 & 0.190 & 0.127 & 0.684 & 0.190 & 0.152 & 0.658 & 0.684 & 0.658 \\
         COCO-2017               & Real      & 10.00 \tiny{$\pm 0.000$} & 9.470 \tiny{$\pm 2.162$} & $\downarrow$ 0.530 & 0.000 & 0.000 & 1.000 & 0.050 & 0.000 & 0.950 & 1.000 & 0.950 \\
         FLICKR-30k              & Real      & 9.950 \tiny{$\pm 0.500$} & 9.460 \tiny{$\pm 1.766$} & $\downarrow$ 0.490 & 0.000 & 0.010 & 0.990 & 0.020 & 0.070 & 0.910 & 0.990 & 0.910 \\
         CIFAR10                 & Real      & 8.380 \tiny{$\pm 2.905$} & 8.180 \tiny{$\pm 2.959$} & $\downarrow$ 0.200 & 0.080 & 0.180 & 0.740 & 0.080 & 0.220 & 0.700 & 0.740 & 0.700 \\
         ImageNet                & Real      & 9.796 \tiny{$\pm 1.428$} & 9.490 \tiny{$\pm 2.179$} & $\downarrow$ 0.306 & 0.020 & 0.000 & 0.980 & 0.051 & 0.000 & 0.949 & 0.980 & 0.949 \\ 
         \hline
         & & \multicolumn{11}{c}{Salesforce/BLIP2-6.7B} \\
         \hline
         SD3.5-Fantasy           & Gen.      & 6.219 \tiny{$\pm 2.804$} & 6.801 \tiny{$\pm 2.240$} & $\uparrow$   0.582 & 0.195 & 0.003 & 0.801 & 0.111 & 0.003 & 0.886 & 0.801 & 0.886 \\
         SD3.5-Outdoor           & Gen.      & 6.222 \tiny{$\pm 2.315$} & 6.584 \tiny{$\pm 1.907$} & $\uparrow$   0.362 & 0.145 & 0.001 & 0.854 & 0.090 & 0.000 & 0.910 & 0.854 & 0.910 \\
         SD3.5-ImageNet          & Gen.      & 4.189 \tiny{$\pm 3.250$} & 5.200 \tiny{$\pm 3.051$} & $\uparrow$   1.011 & 0.484 & 0.000 & 0.516 & 0.326 & 0.000 & 0.674 & 0.516 & 0.674 \\
         \textit{SD1.5}-ImageNet & Gen.      & 5.472 \tiny{$\pm 3.047$} & 6.036 \tiny{$\pm 2.958$} & $\uparrow$   0.564 & 0.284 & 0.016 & 0.700 & 0.228 & 0.012 & 0.760 & 0.700 & 0.760 \\
         CIFAKE                  & Gen.      & 2.770 \tiny{$\pm 3.516$} & 2.840 \tiny{$\pm 3.550$} & $\uparrow$   0.070 & 0.700 & 0.000 & 0.300 & 0.690 & 0.000 & 0.310 & 0.300 & 0.310 \\
         GCC                     & Real      & 3.683 \tiny{$\pm 3.735$} & 3.367 \tiny{$\pm 3.517$} & $\downarrow$ 0.317 & 0.567 & 0.017 & 0.417 & 0.583 & 0.033 & 0.383 & 0.417 & 0.383 \\
         COCO-2017               & Real      & 5.860 \tiny{$\pm 2.425$} & 4.450 \tiny{$\pm 3.208$} & $\downarrow$ 1.410 & 0.190 & 0.040 & 0.770 & 0.390 & 0.030 & 0.580 & 0.770 & 0.580 \\
         FLICKR-30k              & Real      & 5.750 \tiny{$\pm 2.611$} & 1.730 \tiny{$\pm 2.964$} & $\downarrow$ 4.020 & 0.220 & 0.000 & 0.780 & 0.770 & 0.010 & 0.220 & 0.780 & 0.230 \\
         CIFAR10                 & Real      & 3.290 \tiny{$\pm 3.537$} & 3.220 \tiny{$\pm 3.448$} & $\downarrow$ 0.070 & 0.610 & 0.000 & 0.390 & 0.610 & 0.000 & 0.390 & 0.390 & 0.390 \\
         ImageNet                & Real      & 5.351 \tiny{$\pm 2.891$} & 2.454 \tiny{$\pm 3.228$} & $\downarrow$ 2.897 & 0.278 & 0.010 & 0.711 & 0.691 & 0.000 & 0.309 & 0.711 & 0.309 \\ 
         \hline
    \end{tabular}}
    \caption{Realness likelihood ablation across different parameter-sized models and architectures.}
    \vspace{-4mm}
    \label{TABLE_realness_ablation}
\end{table*}

At a lower-level, we observe that without any perturbations, the two BLIP-based models naturally struggle in authenticity detection tasks (see $\overline{P_r(I)}$ column in Table \ref{TABLE_realness_ablation}). For the 2.7B-parameter variant, the model over confidently perceives most generated images as real, oftentimes to a higher degree than images with `real' ground truth labels. In comparison, without applying perturbations, the larger BLIP2-6.7B model generalizes around the mid-point ($\overline{P_r(I)} = 5$), which suggests a higher degree of confusion, which can be exploited through perturbations. This further illustrates how deploying VLMs for tasks in which they are not explicitly optimized for \cite{Zhang2024a} can reveal underlying reliability issues. 

For image captioning tasks, we see that model \textit{architecture} has a greater impact than size. This exposes the importance of labeling and biases within the original training data distributions and the applicability of the VLM for image-captioning tasks. As shown in Fig. \ref{FIG_ablation_bars}(b), Qwen-based models consistently report larger changes in $Y_{\text{VLM}}$ drift in comparison to BLIP models, irrespective of model size. The reported $\overline{N_\text{tokens}}$ columns in Tables \ref{TABLE_caption} and \ref{TABLE_captioning_ablation} can be used to justify these observations. 
BLIP models generally produce less-detailed captions than Qwen, which naturally limits the extent of perturbation-induced output drift. Since Qwen models generate longer outputs, they offer more content to manipulate, making them more susceptible to perturbation effects. Nevertheless, our method still induces noticeable drift even in shorter-captioning models like BLIP \cite{Li2023}.

We demonstrate that our black-box approach generalizes across VLMs, revealing their susceptibility to imperceptible perturbations in the spatial frequency domain. The results confirm the model-agnostic nature of our method. While architecture and size influences the extent of impact across tasks (e.g., authenticity detection vs. captioning), our findings expose a critical vulnerability: VLMs often rely more on image features than true semantic understanding, making them prone to exploitation and means of undermining their reliability.

\begin{table}[]
    \centering
    \resizebox{0.8\linewidth}{!}{%
    \begin{tabular}{l|cccc}
         VLM & task & $H\times W$ & $N_{\text{cand.}}$   & Time (s)\\
         \hline
         \multirow{4}{*}{BLIP2-6.7B} & \multirow{2}{*}{authenticity} & 1024$\times$1024 & 10 & \textbf{36.7875 }\\
                                     &                               & 32$\times$32     & 10 & \textbf{34.4554 }\\
                                     \cline{2-5}
                                     & \multirow{2}{*}{captioning}   & 1024$\times$1024 & 10 & \textbf{182.5212} \\
                                     &                               & 32$\times$32     & 10 & \textbf{229.4264} \\
         \hline
         \multirow{4}{*}{BLIP2-2.7B} & \multirow{2}{*}{authenticity} & 1024$\times$1024 & 10 & 3.4266 \\
                                     &                               & 32$\times$32     & 10 & 0.6550 \\
                                     \cline{2-5}
                                     & \multirow{2}{*}{captioning}   & 1024$\times$1024 & 10 & 4.2586 \\
                                     &                               & 32$\times$32     & 10 & 1.7553 \\
         \hline
         \multirow{4}{*}{Qwen2-7B  } & \multirow{2}{*}{authenticity} & 1024$\times$1024 & 10 & 7.4968 \\
                                     &                               & 32$\times$32     & 10 & 0.6214 \\
                                     \cline{2-5}
                                     & \multirow{2}{*}{captioning}   & 1024$\times$1024 & 10 & 10.9727 \\
                                     &                               & 32$\times$32     & 10 & 2.2303 \\
         \hline
         \multirow{4}{*}{Qwen2.5-3B} & \multirow{2}{*}{authenticity} & 1024$\times$1024 & 10 & 7.4968 \\
                                     &                               & 32$\times$32     & 10 & 0.5962 \\
                                     \cline{2-5}
                                     & \multirow{2}{*}{captioning}   & 1024$\times$1024 & 10 & 10.3863 \\
                                     &                               & 32$\times$32     & 10 & 2.6304 \\
         \hline
         \multirow{4}{*}{Qwen2-2B  } & \multirow{2}{*}{authenticity} & 1024$\times$1024 & 10 & 6.2504 \\
                                     &                               & 32$\times$32     & 10 & 0.5912 \\
                                     \cline{2-5}
                                     & \multirow{2}{*}{captioning}   & 1024$\times$1024 & 10 & 8.0153 \\
                                     &                               & 32$\times$32     & 10 & 1.7753 \\
        \hline
    \end{tabular}}
    \caption{Time taken for each model to do one iteration, searching $N$ candidate perturbations for the next-best perturbed image based on guidance objective.}
    \vspace{-4mm}
    \label{TABLE_times}
\end{table}

\section{Limitations}
Generating optimal perturbations is computationally expensive, especially for image captioning, which requires loading a large additional model during the optimization process. Larger-parameter VLMs and the underlying architecture also has an impact.
We report representative inference time results in Table \ref{TABLE_times} which identifies how evaluation times change \textit{w.r.t.} different models, when exposed to the same test conditions and hyper-parameters. Hardware and resource constraints limit experimenting on very large models and conducting evaluations using cloud-compute resources would incur significant costs.

We acknowledge that this work has harmful implications if exploited with ill-intent. The goal of this work is to identify these easily -exploitable shortcomings in VLMs, with an aim to push research toward addressing these gaps - which we will explore in future works. The limit-imposing and pay-walled nature of enterprise VLM solutions (e.g. Claude, Gemini, GPT) hinders the amount of extensive evaluations that could be done on these widely-popular models. This also justifies deploying publicly-available open source models like Qwen \cite{Bai2023} and BLIP \cite{Li2023} in black-box setups.

\section{Conclusion}

VLMs are increasingly deployed for their perceived reasoning and understanding of image content. We show that, under black-box constraints, their behavior in authenticity detection and automated captioning can be adversarially manipulated. By targeting specific spatial frequency bands, VLM outputs can be misled without introducing perceptible artifacts. Our method imperceptibly perturbs images through task- and decision-guided spatial frequency transformations, and consistently undermines VLM reliability for both real and generated image inputs. This reveals a vulnerability to intrinsic frequency-based cues in VLM reasoning and a lack of sophisticated semantic understanding. As VLMs are proposed as foundational backbones, understanding and mitigating their exploitable cues is essential.


\section{Acknowledgments}
This research and Dr. Jordan Vice are supported by the NISDRG project \#20100007, funded by the Australian Government. Dr. Naveed Akhtar is a recipient of the ARC Discovery Early Career Researcher Award (project \#DE230101058), funded by the Australian Government. Professor Ajmal Mian is the recipient of an ARC Future Fellowship Award (project \#FT210100268) funded by the Australian Government.
\bibliographystyle{IEEEtran}
\bibliography{main}

\end{document}